\documentclass[10pt,journal,cspaper,compsoc]{IEEEtran}

\usepackage{color}
\usepackage{graphicx}
\usepackage{amsmath,amssymb}
\usepackage{times}
\usepackage{multirow}
\usepackage[caption=false]{subfig}
\usepackage{ragged2e}
\usepackage{multirow,array}
\usepackage{cite}
\usepackage{rotating}
\usepackage{multicol,multirow}
\usepackage{overpic}
\usepackage{contour}
\usepackage{booktabs}       

\usepackage{xcolor}
\usepackage{epsfig}
\usepackage{graphicx}
\usepackage{soul}

\usepackage[colorlinks,bookmarksnumbered,bookmarksopen,citecolor=black]{hyperref}
\usepackage{silence}
\usepackage{pifont}
\newcommand{\cmark}{\ding{51}}%
\newcommand{\xmark}{\ding{55}}%

\graphicspath{{./imgs/}{./photos/}}

\def\eg{\emph{e.g.}}
\def\ie{\emph{i.e.}}
\def\etal{et al.}

\def\wrt{\emph{w.r.t.~}}

\renewcommand{\paragraph}[1]{\vspace{3mm} \noindent \textbf{#1}}

\newcommand{\figref}[1]{Fig.~\ref{#1}}
\newcommand{\tabref}[1]{Tab.~\ref{#1}}
\newcommand{\equref}[1]{Eqn.~\ref{#1}}
\newcommand{\secref}[1]{Sec.~\ref{#1}}

\begin{document}

\title{Auto-Rectify Network for Unsupervised Indoor Depth Estimation}

\author{Jia-Wang Bian, Huangying Zhan, Naiyan Wang, Tat-Jun Chin, Chunhua Shen, Ian Reid
\IEEEcompsocitemizethanks{
  \IEEEcompsocthanksitem J.-W. Bian, H. Zhan, T.-J. Chin, C. Shen, and I. Reid are with 
    The University of Adelaide and Australian Centre for Robotic Vision, Australia;
  \IEEEcompsocthanksitem N. Wang is with TuSimple, China.
}
}

\markboth{IEEE Transactions on Pattern Analysis and Machine Intelligence Submission}%
{Bian \MakeLowercase{\textit{et al.}}: Auto-Rectify Network for Unsupervised Depth Learning from Motion-Complex Video}


\IEEEtitleabstractindextext{%
\begin{abstract}
\justifying
Single-View depth estimation using the CNNs trained from unlabelled videos has shown significant promise.
However, excellent results have mostly been obtained in street-scene driving scenarios, 
and such methods often fail in other settings, particularly indoor videos taken by handheld devices.
In this work, we establish that the complex ego-motions exhibited in handheld settings are a critical obstacle for learning depth.
Our fundamental analysis suggests that the rotation behaves as noise during training, as opposed to the translation (baseline) which provides supervision signals.
To address the challenge, we propose a data pre-processing method that rectifies training images by removing their relative rotations for effective learning.
The significantly improved performance validates our motivation.
Towards end-to-end learning without requiring pre-processing,
we propose an Auto-Rectify Network with novel loss functions,
which can automatically learn to rectify images during training.
Consequently, our results outperform the previous unsupervised SOTA method by a large margin
on the
challenging NYUv2 dataset.
We also demonstrate the generalization of our trained model in ScanNet and Make3D,
and the universality of our proposed learning method
on 
7-Scenes and KITTI datasets.
\end{abstract}

}
\maketitle
\IEEEraisesectionheading{\section{Introduction}\label{sec:intro}}


\IEEEPARstart{I}{nferring} 3D geometry from 2D images is a long-standing problem in robotics and computer vision.
Depending on the specific use case, it is usually solved by Structure-from-Motion~\cite{schonberger2016structure} or Visual SLAM~\cite{davison2007monoslam, newcombe2011dtam, mur2015orb}.
Underpinning these traditional pipelines is searching for correspondences~\cite{lowe2004distinctive,Bian2019Gms} across multiple images and triangulating them via epipolar geometry~\cite{Zhang1998,hartley2003multiple,bian2019bench} to obtain 3D points.
Following the growth of deep learning-based approaches, 
Eigen~\etal~\cite{eigen2014depth} show that the depth map can be inferred from a single color image by a CNN,
which is trained with the ground-truth depth supervisions captured by range sensors. 
Subsequently a series of supervised methods~\cite{liu2016learning, eigen2015predicting, chakrabarti2016depth, laina2016deeper, li2017two, fu2018deep, Yin2019enforcing} have been proposed and the accuracy of estimated depth is progressively improved.

Based on epipolar geometry~\cite{Zhang1998,hartley2003multiple,bian2019bench}, learning depth without requiring the ground-truth supervision has been explored.
Garg~\etal~\cite{garg2016unsupervised} showed that the single-view depth CNN can be trained from stereo image pairs with known baseline via photometric loss.
Zhou~\etal~\cite{zhou2017unsupervised} further explored the unsupervised framework and proposed to train the depth CNN from unlabelled videos. 
They additionally introduced a Pose CNN to estimate the relative camera pose between consecutive frames, and 
the photometric loss remains as the main supervision signal.
Following that, a number of of unsupervised methods have been proposed, which can be categorised into stereo-based~\cite{godard2017unsupervised, zhan2018unsupervised, zhan2019self, watson2019self} and video-based~\cite{Wang2018CVPR, mahjourian2018unsupervised, yin2018geonet, zou2018df, ranjan2019cc, monodepth2, gordon2019depth, chen2019self, bian2021ijcv}, according to the type of training data.
Our work follows the latter paradigm, since unlabelled videos are easier to obtain in real-world scenarios.

Unsupervised methods have shown promising results in driving scenes, \eg, KITTI~\cite{Geiger2013IJRR} and Cityscapes~\cite{Cordts2016Cityscapes}.
However, as reported in~\cite{Zhou_2019_ICCV}, they usually fail in generic scenarios such as the indoor scenes in NYUv2 dataset~\cite{silberman2012indoor}.
For example, GeoNet~\cite{yin2018geonet}, one of the state-of-the-art methods in KITTI, is unable to obtain reasonable results in NYUv2.
To this end, \cite{Zhou_2019_ICCV} proposes to use optical flow as the supervision signal to train the depth CNN,
and \cite{zhao2020towards} also uses flow with epipolar geometry to estimate ego-motion to replace the Pose CNN.
However, the reported depth accuracy \cite{zhao2020towards} is still limited, \ie, 0.189 in terms of \emph{AbsRel}---see also qualitative results in \figref{fig:show}.

Our work investigates the fundamental reasons behind poor results of unsupervised depth learning in indoor scenes.
In addition to the usual challenges such as non-Lambertian surfaces and low-texture scenes, we identify the camera motion profile in the training videos as a critical factor that affects the training process. To develop this insight, we conduct an in-depth analysis of the effects of camera pose to current unsupervised depth learning framework. Our analysis shows that
(i) fundamentally the camera rotation behaves as noise to training, while the translation contributes effective gradients; (ii) the rotation component dominates the motions in indoor videos captured using handheld cameras, while the opposite is true in autonomous driving scenarios.

Inspired by image rectification~\cite{fusiello2000compact} which has been used a standard pre-processing for the stereo matching task~\cite{hirschmuller2007stereo},
we propose to weakly rectify video frames for the effective depth learning.
To be specific, we compute the relative rotation between image pairs and warp them to their common image plane,
which results in rectified pairs that have no relative rotational motions,
and we use these pairs for unsupervised depth learning.
This is achieved by leveraging the traditional feature matching~\cite{lowe2004distinctive, Bian2019Gms} and epipolar geometry~\cite{nister2004efficient, hartley2003multiple},
while no ground truth data is required.
With our proposed data pre-processing, we demonstrate that existing state-of-the-art (SOTA) unsupervised methods~\cite{monodepth2,bian2021ijcv} can be trained well in the challenging indoor NYUv2 dataset~\cite{silberman2012indoor}. 
Consequently, our results outperform the unsupervised SOTA~\cite{zhao2020towards} by a large margin. 

Furthermore, towards end-to-end learning without requiring a manual pre-processing,
we propose an Auto-Rectify Network (ARN),
which draws philosophies from Spatial Transformer Network~\cite{jaderberg2015stn} and elegantly incorporates the proposed weak rectification into the modern deep learning framework.
The ARN, along with our proposed novel loss functions, can automatically learn to rectify images during the unsupervised training.
The obtained results are comparable to training models on the pre-processed data.
We make comprehensive ablation studies on the proposed methods,
and validates the generalization of our trained model and learning method in several standard datasets, including ScanNet~\cite{dai2017scannet}, 7-Scenes~\cite{shotton2013scene}, and KITTI~\cite{Geiger2013IJRR}.

To summarize, our main contributions include:
\begin{itemize}
    \item We theoretically analyze the relation between depth and two motion components in the warping. 
    Along with the experimental analysis of the camera motion distribution in different scenarios,
    we answer the question why it is so challenging to train unsupervised depth CNNs from indoor videos captured by handheld cameras.
    \item We propose a data pre-processing method to address the complex motion, which significantly boosts the learned depth accuracy and validates our motivation. 
    \item We propose an Auto-Rectify Network with novel loss functions towards end-to-end learning, resulting in an improved unsupervised depth learning framework.
\end{itemize}

\section{Related Work}

This paper address the challenges of unsupervised learning of monocular depth from motion-complex videos, which are often captured by a hand-held camera.
Based on our findings that the rotation between image pairs makes depth learning more challenging,
we propose two image rectification methods for removing the rotation by either data pre-processing or end-to-end learning.
The related work is discussed in the following paragraphs. 

\paragraph{Unsupervised Video Depth Estimation.}
Zhou~\etal~\cite{zhou2017unsupervised} propose the first video-based unsupervised learning method for monocular depth estimation.
They rely on the color consistency between consecutive frames and the underlying 3D geometry constraint to train both depth and pose networks jointly.
Following this seminal work, \cite{yin2018geonet, zou2018df, ranjan2019cc, casser2019struct2depth, bian2021ijcv, chen2019self, lee2021learning, monodepth2} propose different contributions to address the moving object issue in the dynamic scenes.
\cite{mahjourian2018unsupervised, Wang2018CVPR} propose additional geometric optimization to improve the pose estimation,
and \cite{packnet} proposes a better depth network architecture.
The significantly improved results are presented,
however, these methods only demonstrate high-quality depth estimation results in driving videos, \eg, KITTI~\cite{Geiger2013IJRR} or Cityscapes~\cite{Cordts2016Cityscapes} datasets.
As pointed by \cite{Zhou_2019_ICCV}, many state-of-the-art methods such as GeoNet~\cite{yin2018geonet} fail to get reasonable results in NYUv2 indoor dataset~\cite{silberman2012indoor}.
We experiment with Monodepth2~\cite{monodepth2},
which sometimes also meets this issue. 
Through careful investigation, we identify the complex camera pose in the hand-held camera setting as the main challenge,
and we address the challenge in this paper.

Before us, Zhou~\etal~\cite{Zhou_2019_ICCV} and Zhao~\etal~\cite{zhao2020towards} have experimented in the indoor NYUv2 dataset.
Specifically, \cite{Zhou_2019_ICCV} estimates the optical flow between image pairs, and then uses the flow instead of photometric loss to supervise the depth and pose based warping.
\cite{Zhou_2019_ICCV} also removes purely rotated image pairs from training data,
while the rotation in the remaining training image pairs are not addressed.
We instead deal with the rotation in all image pairs automatically in this paper.
\cite{zhao2020towards} replaces the pose network with an optical flow assisted geometric module for pose estimation.
However, they did not realize the rotation issue in the hand-held camera setting, and the performance is limited.
Besides, a recent work~\cite{IndoorSfMLearner} proposes to address the low-texture issue in the indoor scene by using patch-match and plane regularization.
Their contributions are complementary to ours,
and the results show that our method achievers a higher performance.

\paragraph{Image Rectification.}
Rectifying images for better depth estimation is not new in the computer vision community.
For example, stereo rectification~\cite{fusiello2000compact} makes left and right images captured by a stereo camera to be row-to-row aligned,
which is a widely used pre-processing for stereo matching~\cite{hirschmuller2007stereo}.
Besides, in the classic multi-view stereo system~\cite{schoenberger2016mvs},
images are also rectified before depth triangulation and optimization.
We borrow this idea from classic geometry algorithms and integrate it into the deep learning based framework for monocular depth learning from motion-complex videos.
A similar idea is used in \cite{Do2020SurfaceNormal} to learn the surface normal of tiled images.
Moreover, our method is related to Spatial Transform Networks~\cite{jaderberg2015spatial} and Homography estimation~\cite{detone2016deep}.
The former learns the image transform parameters and proposes a differentiable image warping,
and the latter also learns the image rotation.
The detailed differences are discussed in \secref{sec:arn-design}.


\section{Problem Analysis}

We first overview the unsupervised framework for depth learning.
Then, we revisit the depth and ego-motion based image warping and demonstrate the relationship between depth and decomposed motions.
Finally, we compare the ego-motion statistics in different scenarios to verify the impact of ego-motion on depth learning.

\subsection{Unsupervised depth learning from video}\label{sec:sc-overview}

\begin{figure*}[t]
\centering
\includegraphics[width=0.98\linewidth]{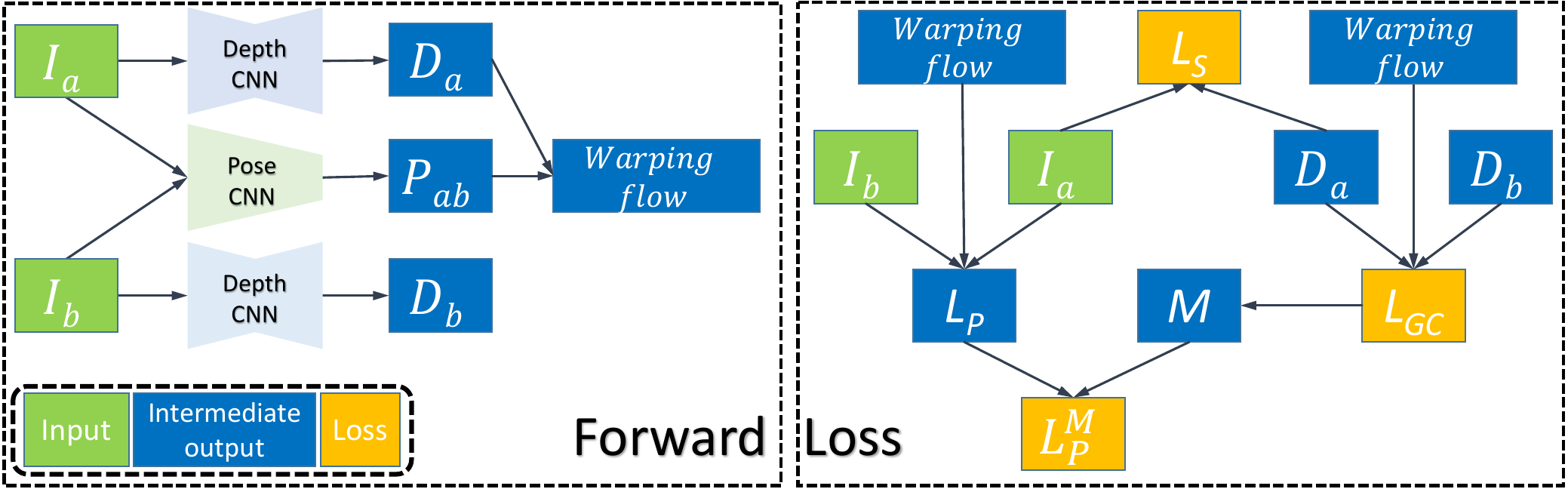} 
\caption{Overview of SC-Depth~\cite{bian2021ijcv}.
Firstly, in the forward pass, training images ($I_a$, $I_b$) are passed into the network to predict depth maps ($D_a$, $D_b$) and relative camera pose $P_{ab}$.
With $D_a$ and $P_{ab}$, we obtain the warping flow between two views according to \equref{eqn:full_tansformation}.
Secondly, given the warping flow, the photometric loss $L_P$ and the geometry consistency loss $L_{GC}$ are computed.
Also, the weighting mask $M$ is derived from $L_{GC}$ and applied over $L_P$ to handle dynamics and occlusions.
Moreover, an edge-aware smoothness loss $L_S$ is used to regularize the predicted depth map.
Here we use this system as our baseline for problem analysis and illustrate our proposed components in \figref{fig:arn}.
}
\label{fig:sc-overview}
\end{figure*}

Following SfMLearner~\cite{zhou2017unsupervised}, plenty of unsupervised depth estimation methods have been proposed. The SC-Depth~\cite{bian2021ijcv}, which is the current SOTA framework, additionally constrains the geometry consistency over \cite{zhou2017unsupervised},
leading to more accurate and scale-consistent results.
In this paper, we use SC-Depth as our baseline,
and we overview its pipeline in \figref{fig:sc-overview}.

\paragraph{Forward.}
A training image pair ($I_a$, $I_b$) is first passed into
a weight-shared depth CNN to obtain the depth maps ($D_a$, $D_b$), respectively.
Then, the pose CNN takes the concatenation of two images as input and predicts their 6D relative camera pose $P_{ab}$.
With the predicted depth $D_a$ and pose $P_{ab}$,
the warping flow between two images is generated according to \secref{sec:proof}.

\paragraph{Loss.}
First, the main supervision signal is the photometric loss $L_P$.
It calculates the color difference in each pixel between $I_a$ with its warped position on $I_b$ using a \emph{differentiable bilinear interpolation}~\cite{jaderberg2015stn}.
Second, depth maps are regularized by the geometric inconsistency loss $L_{GC}$, where it enforces the consistency of predicted depths between different frames.
Besides, a weighting mask $M$ is derived from $L_{GC}$ to handle dynamics and occlusions, which is applied on $L_P$ to obtain the weighted $L_{P}^M$.
Third, depth maps are also regularized by a smoothness loss $L_S$,
which ensures that depth smoothness is guided by the edge of images.
Overall, the objective function is:
\begin{equation}\label{eqn:loss}
L = \alpha L_{P}^M + \beta L_{S} + \gamma L_{GC},
\end{equation}
where $\alpha$, $\beta$, and $\gamma$ are hyper-parameters to balance different losses.

\subsection{Depth and camera pose based image warping}\label{sec:proof}
The image warping builds the link between networks and losses during training,
\ie, the warping flow is generated by network predictions (depth and camera motion) in forward pass,
and the gradients are back-propagated from the losses via the warping flow to networks.
Therefore, we investigate the warping to analyze the camera pose effects on the depth learning, 
which avoids involving image content factors, such as illumination changes and low-texture scenes.

\paragraph{Full transformation.}
The camera pose is composed of rotation and translation components.
A point ($u_1$, $v_1$) in the first image is reprojected to ($u_2$, $v_2$) in the second image, which satisfies the following relationship:
\begin{equation}\label{eqn:full_tansformation}
{\bf K}^{-1} \left(d_2 \begin{bmatrix} u_2 \\ v_2 \\ 1\end{bmatrix}\right)
= {\bf R} {\bf K}^{-1} \left(d_1 \begin{bmatrix} u_1 \\ v_1 \\ 1\end{bmatrix}\right) + {\bf t},
\end{equation}
where $d_{i}$ is the depth of this point in two images and ${\bf K}$ is the 3x3 camera intrinsic matrix.
${\bf R}$ is a 3x3 rotation matrix and ${\bf t}$ is a 3x1 translation vector.
We decompose the full warping flow and discuss each component below.
 
\paragraph{Pure-rotation transformation.}
If two images are related by a pure-rotation transformation (\ie, ${\bf t} = {\bf 0}$),
based on \equref{eqn:full_tansformation},
the warping satisfies:
\begin{equation}
 d_2
 \begin{bmatrix}  u_{2} \\ v_{2}\\ 1 \end{bmatrix}
 = {\bf K} {\bf R} {\bf K}^{-1} \left(d_1 \begin{bmatrix}  u_{1} \\ v_{1}\\ 1 \end{bmatrix}\right),
 \end{equation}
where $[{\bf K}{\bf R}{\bf K}^{-1}]$ is as known as the \emph{homography matrix} ${\bf H}$~\cite{hartley2003multiple},
and we have

\begin{equation}
\begin{bmatrix}
u_{2} \\ v_{2}\\ 1 
\end{bmatrix} 
= 
\frac{d_1}{d_2} \bf H 
\begin{bmatrix}  u_{1} \\ v_{1}\\ 1 \end{bmatrix}
= c \begin{bmatrix}  
h_{11} & h_{12} & h_{13}  \\
h_{21} & h_{22} & h_{23}  \\
h_{31} & h_{32} & h_{33}
\end{bmatrix}
\begin{bmatrix}  u_{1} \\ v_{1}\\ 1 \end{bmatrix},
\end{equation}
where $c=\frac{d1}{d2}$, standing for the depth relation between two views, is determined by the third row of the above equation,
\ie, $c = 1 / (h_{31} u_1 + h_{32} v_1 + h_{33})$.
It indicates that we can obtain ($u_2$, $v_2$) without $d_1$.
Specifically, solving the above equation, we have
\begin{equation}\label{eqn:homography}
\begin{cases}
u_2 = (h_{11} u_1 + h_{12} v_1 + h_{13}) / (h_{31} u_1 + h_{32} v_1 + h_{33})   \\
v_2 = (h_{21} u_1 + h_{22} v_1 + h_{23}) / (h_{31} u_1 + h_{32} v_1 + h_{33}) .
\end{cases}
\end{equation}

This demonstrates that the rotational flow in image warping is independent to the depth,
and it is only determined by ${\bf K}$ and ${\bf R}$.
Consequently, the rotational motion in image pairs cannot contribute effective gradients to supervise the depth CNN during training,
even when it is correctly estimated.
More importantly, if the estimated rotation is inaccurate,
noisy gradients will arise and harm the depth CNN in backpropagation.
Therefore, we conclude that the rotational motion behaves as the noise to unsupervised depth learning.

\paragraph{Pure-translation transformation.}
A pure-translation transformation means that ${\bf R}$ is an identity matrix in \equref{eqn:full_tansformation}.
Then we have
\begin{equation} \small
d_2 
\begin{bmatrix}  u_{2} \\ v_{2}\\ 1 \end{bmatrix}
= d_1 \begin{bmatrix}  u_{1} \\ v_{1} \\ 1 \end{bmatrix} + {\bf K} {\bf t}
= d_1 \begin{bmatrix}  u_{1} \\ v_{1} \\ 1 \end{bmatrix} + 
\begin{bmatrix}  f_x & 0 & c_x \\
0 & f_y & c_y \\
0 & 0 & 1
\end{bmatrix}
\begin{bmatrix}  t_1 \\ t_2 \\ t_3 \end{bmatrix},
\end{equation}
where ($f_x$ $f_y$) are camera focal lengths, and ($c_x$, $c_y$) are principal point offsets.
Solving the above equation, we have
\begin{equation}\label{eqn:translation}
\begin{cases}
d_2 u_2 = d_1 u_1 + f_x t_1 + c_x t_3 \\
d_2 v_2 = d_1 v_1 + f_y t_2 + c_y t_3 \\
d_2  = d_1 + t_3
\end{cases}
\begin{cases}
u_2 = \frac{d_1 u_1 + f_x t_1 + c_x t_3}{d_1 + t_3} \\ \\ 
v_2 = \frac{d_1 v_1 + f_y t_2 + c_y t_3}{d_1 + t_3}. \\  
\end{cases}
\end{equation}
It shows that the translation vector ${\bf t}$ is coupled with the depth $d_1$ during the warping from ($u_1$, $v_1$) to ($u_2$, $v_2$).
This builds the link between the depth CNN and the warping,
so that gradients from the photometric loss can flow to the depth CNN via the warping.
Therefore, we conclude that the translational motion provides effective supervision signals to depth learning.

\subsection{Distribution of decomposed camera motions}\label{sec:pose-effects}

\begin{figure}[t]
\centering
\footnotesize
KITTI~\cite{Geiger2013IJRR} (R=$0.25^{\circ}$, T=$0.99m$) \\
\includegraphics[width=0.7\linewidth]{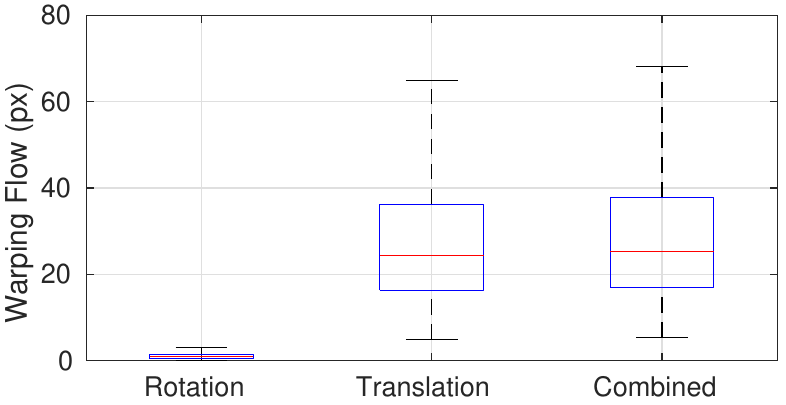} \\ \vspace{1mm}
Original NYUv2~\cite{silberman2012indoor} (R=$2.28^{\circ}$, T=$0.05m$) \\
\includegraphics[width=0.7\linewidth]{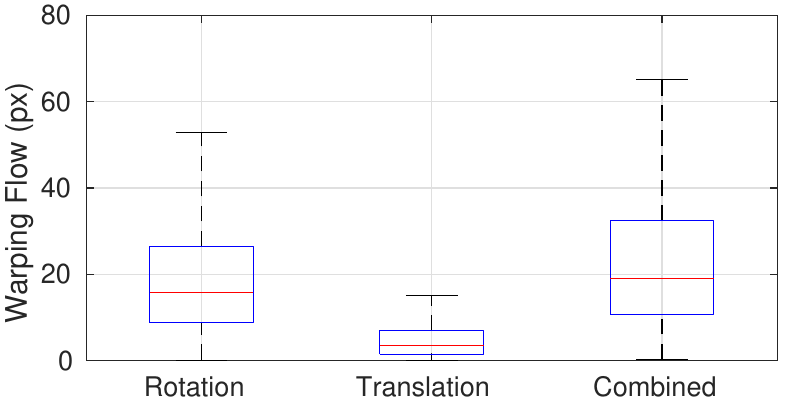} \\ \vspace{1mm}
Rectified NYUv2 (R=$0.68^{\circ}$, T=$0.24m$) \\
\includegraphics[width=0.7\linewidth]{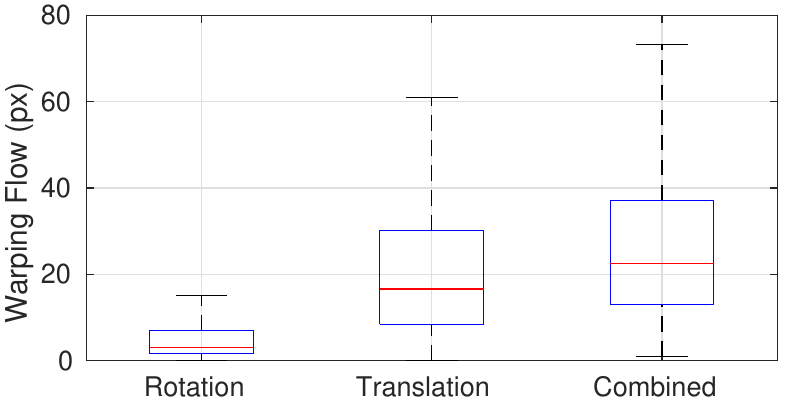} \\
\caption{Distribution of inter-frame camera motions and warping flows.
"Rectified" stands for the proposed pre-processing method described in \secref{sec:method}.
In each figure, the first row shows the averaged magnitude of camera poses, \ie, \textbf{R} for rotation and \textbf{T} for translation,
and the plot shows the distribution of decomposed warping flow magnitudes (px) of randomly sampled points.
}
\label{fig:pose-statistics}
\end{figure}

\paragraph{Inter-frame camera motions and warping flows.}
\figref{fig:pose-statistics} shows the camera motion statistics on KITTI~\cite{Geiger2013IJRR} and NYUv2~\cite{silberman2012indoor} datasets.
KITTI is pre-processed by removing static images, as done in~\cite{zhou2017unsupervised, bian2021ijcv}.
We pick one image of every 10 frames in NYUv2, which is denoted as \emph{Original NYUv2}.
Then we apply the proposed pre-processing (\secref{sec:method}) to obtain \emph{Rectified NYUv2}.
For all datasets, we compare the decomposed camera pose of their training image pairs \wrt the absolute magnitude and inter-frame warping flow\footnote{We first compute the rotational flow using \equref{eqn:homography}, and then we get the translational flow by subtracting rotational flow from the overall warping flow. Here, the translational flow is also called \emph{residual parallax} in \cite{Li2020MannequinChallengeLT}, where it is used to compute depth from correspondences and relative camera poses.}.
Specifically, we compute the averaged warping flow of randomly sample points in the first image using the ground-truth depth and pose.
For each point ($u_1$, $v_1$) that is warped to ($u_2$, $v_2$), the flow magnitude is $\sqrt{(u_2-u_1)^2 + (v_2-v_1)^2}$.
\figref{fig:pose-statistics} shows that the rotational flows dominates the flows in \emph{Original NYUv2} but it is opposite in KITTI.
Along with the conclusion in \secref{sec:proof} that the depth is supervised by the translation while the rotation behaves as the noise,
this answers the question why unsupervised depth learning methods that obtain state-of-the-art results in driving scenes often fail in indoor videos.
Besides, the results on \emph{Rectified NYUv2} demonstrate that our proposed data pre-processing can effectively address the issue.

\section{Proposed data processing}\label{sec:method}

The above analysis suggests that unsupervised depth learning favours image pairs those have small rotational and moderate translational motions for training.
However, unlike driving sequences, videos captured by handheld cameras tend to have significant rotational motions, as shown in \figref{fig:pose-statistics}.
In this section, we describe the proposed method to rectify images for effective learning in \secref{sec:weak-rectify},
and discuss the method in~\secref{sec:discussion-wr}.

\subsection{Weak Rectification}\label{sec:weak-rectify}

\paragraph{Sampling image pair candidates.}
Given a high frame rate video, \eg, $30$fps,
we first downsample it temporally to remove redundant images, \ie, extract one key (1st) frame from every $m$ frames.
Then, instead of only considering adjacent frames as a pair, 
we pair up each image with its following $k$ frames as pair candidates.
Here, we use $m=10$ and $k=10$ in NYUv2~\cite{silberman2012indoor}.

\paragraph{Two-view relative pose estimation.}
For each pair candidate, we first generate correspondences across images by using SIFT~\cite{lowe2004distinctive} features,
and we then apply the \emph{ratio test}~\cite{lowe2004distinctive} with GMS~\cite{bian2021ijcv} filter to find good matches.
Second, with the selected correspondences, we estimate the \emph{essential matrix} using the five-point algorithm~\cite{nister2004efficient} within a RANSAC~\cite{fischler1981random} framework.
Then the relative camera pose is decomposed from the essential matrix.
This two-view image matching pipeline has a well-established technique in Computer Vision.
See more technical details in~\cite{hartley2003multiple,bian2019bench}.

\paragraph{3-DoF weak rectification.}
Given an image pair with the pre-computed relative pose,
we warp both images to a common plane using the their relative rotation matrix $\bf R$.
Specifically, (i) we fist convert $\bf R$ to rotation vector $\bf r$ using Rodrigues formula~\cite{trucco1998introductory} to obtain half rotation vectors for two images (\ie, $\frac{\bf r}{2}$ and $-\frac{\bf r}{2}$),
and then we convert them back to rotation matrices $\bf R_1$ and $\bf R_2$.
(ii) Given $\bf R_1$, $\bf R_2$, and camera intrinsic $\bf K$, we warp images to a new common image plane according to \equref{eqn:homography}.
Then in the common plane, we crop their overlapped rectangular regions to obtain the weakly rectified pairs.

\subsection{Discussion of the proposed weak rectification}\label{sec:discussion-wr}

\paragraph{Inspiration from Stereo Rectification~\cite{fusiello2000compact}.}
The image rectification has been standard techniques for pre-processing images in the stereo matching task~\cite{hirschmuller2007stereo},
which simplifies the problem by rectifying images so that corresponding points in two images have identical vertical coordinates.
Drawing inspiration from it, we propose to weakly rectify images for easing the task of unsupervised depth learning from video.
Compared with the standard rectification~\cite{fusiello2000compact},
we only consider the rotation for image warping and deliberately ignore the translation,
which results in weakly rectified pairs that have 3-DoF translational motions,
instead of 1-DoF motions as in~\cite{fusiello2000compact},
due to two reasons:
(i) the standard rectification is used for imaged pairs that are captured by two horizontal cameras,
and using it in video cases, especially the forward-motion pairs, often cause wired results such as extremely deformed images~\cite{fusiello2000compact};
(ii) the rigorous 1-DoF rectification is unnecessary for unsupervised depth learning,
because the motions in all three directions are associated with the depth during image warping, as shown in \equref{eqn:translation}.
An experimental example is that the existing methods~\cite{zhou2017unsupervised,yin2018geonet,ranjan2019cc,monodepth2, bian2021ijcv} that are trained on KITTI videos,
where image pairs have 3-DoF translational motions,
can show comparable results to methods those are trained on KITTI stereo pairs~\cite{garg2016unsupervised, godard2017unsupervised, zhan2018unsupervised}.


\paragraph{How to deal with unstable rectification?}
This is unavoidable due to the nature of geometric methods.
For example, we may obtain significantly deformed images due to the inaccurate rotation estimation,
which could be caused by insufficient correspondence search.
However, this does not matter in our problem,
because our goal is to construct good data for effective training, 
while we do not require that the rectification on all data is perfect.
In practice, we set thresholds to remove invalid pairs,
\eg, based on correspondence numbers and aspect ratios of the rectified images.

\paragraph{Can we do rectification in end-to-end fashion?}
Although using our pre-processed data can lead to better performance (\tabref{tab:nyu}) and faster convergence (\figref{fig:validation-loss}),
the end-to-end learning approach is more favorable.
For example, it benefits post-processing during inference,
as recently discussed in~\cite{chen2019self, luo2020consistent}.
We here propose the data pre-processing solution mainly to validate our motivation,
and we show how to encapsulate this idea into an end-to-end learning framework in the next section.

\begin{figure}[t]
\centering
\includegraphics[width=0.95\linewidth]{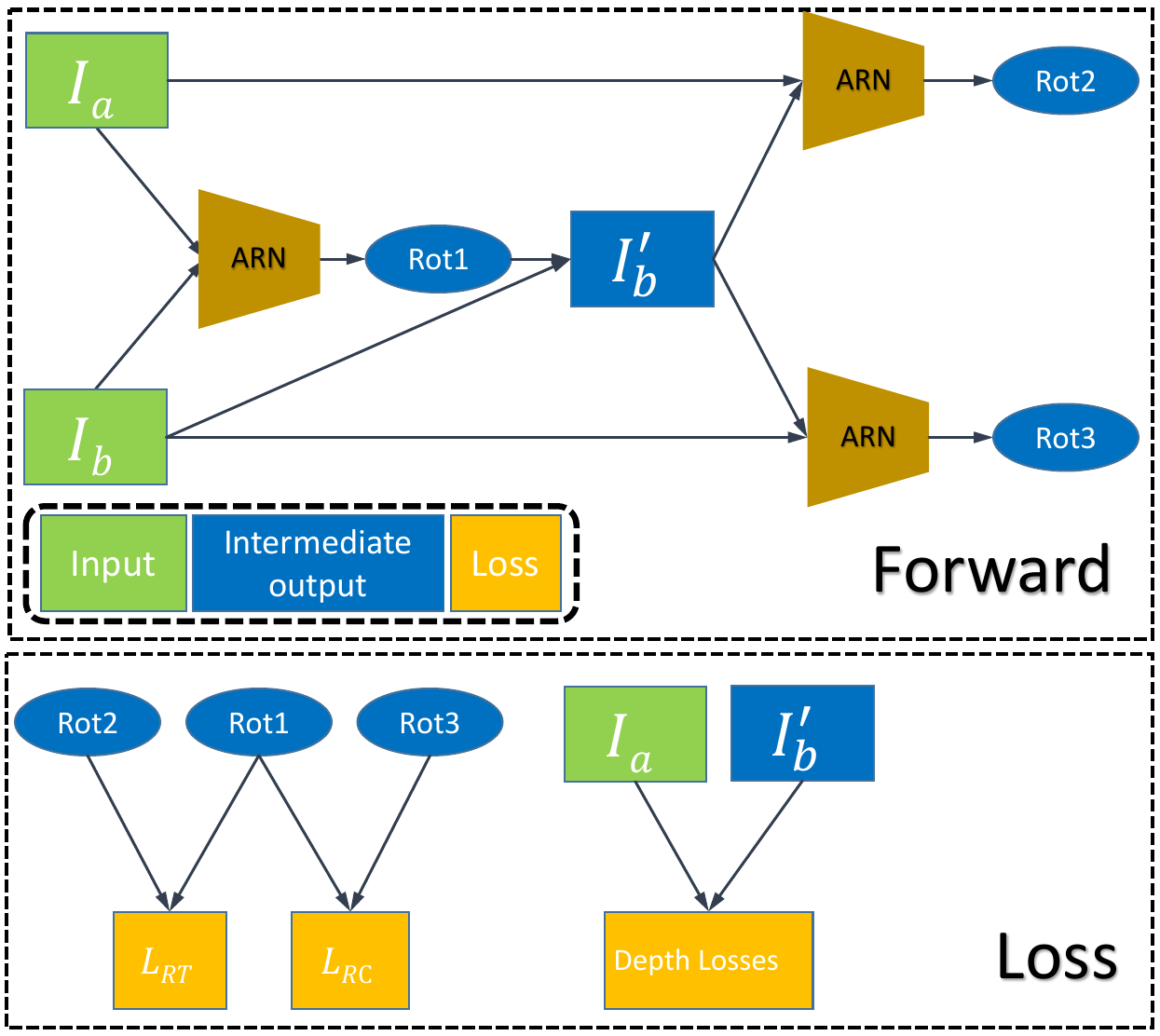} 
\caption{Proposed Auto-Rectify Network (ARN) with loss functions.
We use ARN to predict the relative rotation between two input images ($I_a, I_b$),
and warp $I_b$ to obtain the $I_b'$,
which is supposed to be well-aligned with $I_a$ in terms of rotation.
Then we use the image pair ($I_a, I_b'$) for subsequent depth learning, as described in \figref{fig:sc-overview}.
The proposed loss functions are used to regularize the training of ARN.
}
\label{fig:arn}
\end{figure}

\begin{figure}[ht]
\centering
\includegraphics[width=0.98\linewidth]{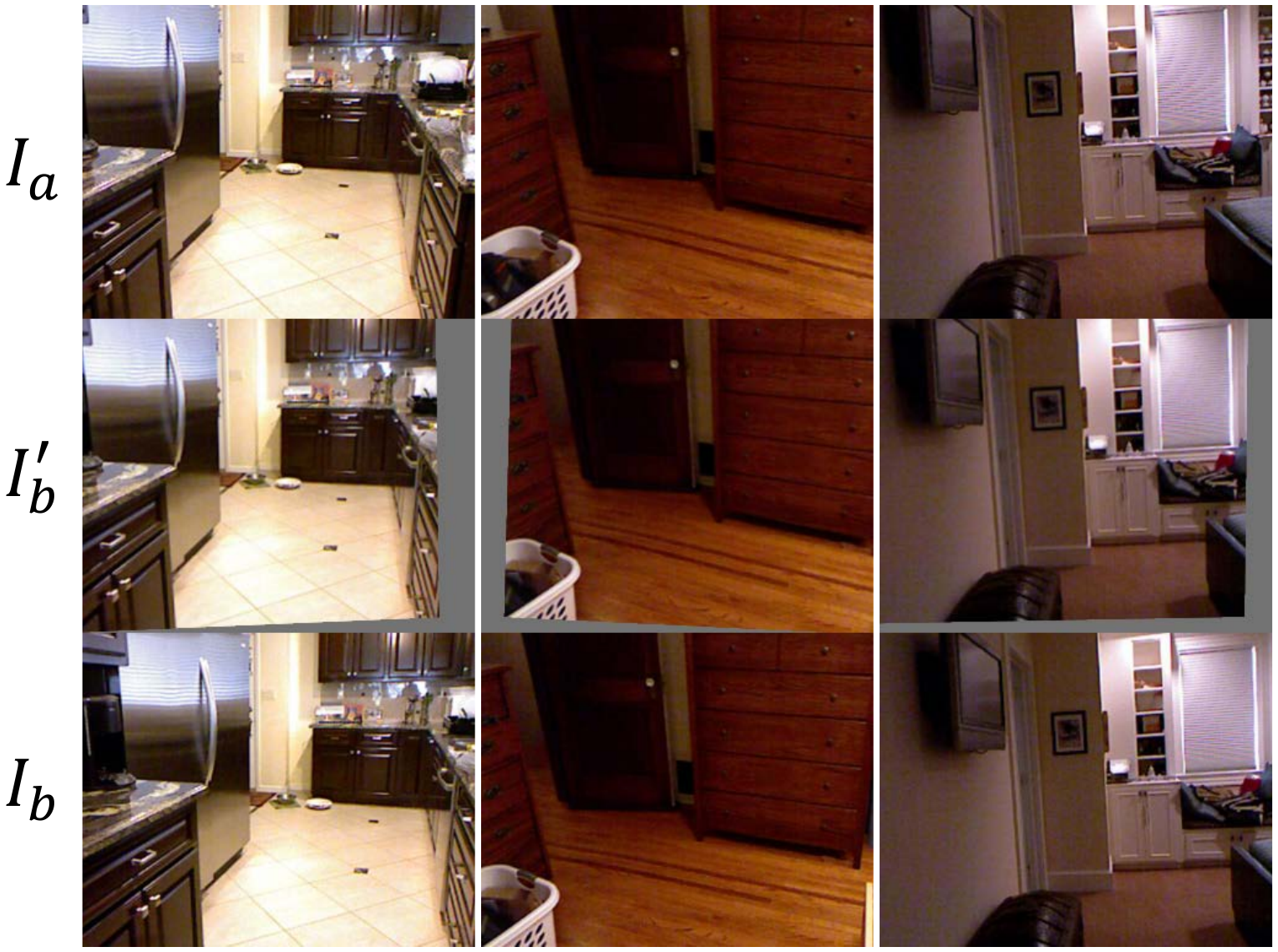}
\caption{Samples of ARN warped results. $I_a$, $I_b$ are input images, and $I_b'$ is the warped image by ARN. Note the grey board of $I_b'$, which stands for the zero-padding in image warping.}
\label{fig:arn-vis}
\end{figure}

\section{Proposed end-to-end method}
In this section,
we present the proposed Auto-Rectify Network (ARN) with novel loss functions in \secref{sec:arn} for end-to-end learning,
and we discuss our design in \secref{sec:arn-design}.

\subsection{Auto-Rectify Network with losses}\label{sec:arn}

\figref{fig:arn} illustrates the proposed methods.
ARN takes a concatenation of two images as input and outputs their 3-DoF relative rotation.
Then we warp $I_b$ to obtain $I_b'$ using the predicted rotation parameters,
which is supposed to be co-planar with $I_a$.
Then we pass the image pair ($I_a$, $I_b'$) to our baseline framework for training.
\figref{fig:arn} shows the pipeline and \figref{fig:arn-vis} shows the visualized results.


\paragraph{Network Architecture.}
The network architectures of ARN is almost the same as Pose CNN, and the difference is that ARN outputs a 3-DoF rotation instead of full 6-DoF pose.
We use ResNet-18~\cite{he2016deep} encoder to extract features from a concatenation of two images,
and regress their rotation parameters using several convolution layers.
In order to enable two frames as input,
we modify the first layer to have six channels instead of three.

\paragraph{Loss Functions.}
First, as we expect that $I_b'$ is well-aligned with $I_a$ \wrt rotation, $Rot2$ is minimized. 
However, this leads to a trivial solution (\ie, the network always outputs zero), 
which inhabits the network ability for learning rotation and thus rectification.
As a regularization, we simultaneously maximize $Rot1$.
This relies in the assumption that the warped pairs are more well-aligned than original pairs.
This allows us to define a Rotation-Triple loss $L_{RT}$.
Formally,
\begin{equation}
    L_{RT} = max(||Rot2||_1 - ||Rot1||_1 + \delta, 0),
\end{equation}
where $\delta$ is the margin, which we empirically set to be $0.1$.
Besides, as $I_b'$ is manually warped from $I_b$ using $Rot1$,
which is the ground truth of $Rot3$,
the prediction of ARN on pair ($I_b, I_b'$).
Therefore, we define an Rotation-Consistent loss $L_{RC}$:
\begin{equation}
    L_{RC} = ||Rot3 - Rot1||_1,
    \label{eqn:l_rc}
\end{equation}
where $||\cdot||_1$ stands for the $L_1$ loss.
We validate the effects of both losses in \tabref{tab:ablation-loss}.

\subsection{Discussion of the proposed ARN}\label{sec:arn-design}

\paragraph{Inspiration from Spatial Transformer Network~\cite{jaderberg2015stn}}.
The STN is proposed to warp a single image (or its feature map) to the canonical view using the automatically learned spatial transformation parameters,
and this has shown advantages to supervised learning tasks, \eg, image classification.
In this paper, drawing inspiration from STN, we propose ARN to learn the relative rotation between two images and rectify images by warping the second image using the learned rotation parameters to a new view, which is supposed to be better aligned with the first image.
We demonstrate that using the rectified pairs leads to significantly better performance than original pairs for unsupervised depth learning in motion-complex dataset---see \tabref{tab:nyu}.

\paragraph{Relation to learning homography.}
Although both methods estimate a relative rotation, homography estimators solve rotation of purely rotated image pairs for accurate registration,
while the goal of our proposed ARN is to rectify image pairs with general motions.
In the special case where input images are related by only rotational motions,
our ARN is equivalent to the homography estimator.
Therefore, we propose the $L_{RC}$ (\equref{eqn:l_rc}) for supervising the network training in such case.
Specifically, $Rot3$ is the prediction of ARN on the purely rotated pairs ($I_b, I_b'$),
which is manually rotated by $Rot1$,
so we minimize the difference between $Rot3$ and $Rot1$ to encourage 
accurate rotation predictions.

\paragraph{How to deal with purely rotated pairs?}
As \equref{eqn:homography} shows that the rotation behaves as noises during training,
so that purely rotated pairs are invalid for training the depth CNN.
Previous methods~\cite{Zhou_2019_ICCV, zhao2020towards} manually remove that from the dataset.
However, we show that in our end-to-end learning framework,
the purely rotated images could be automatically masked out.
Specifically, for the image pair ($I_a, I_b$) that is related by purely rotational motions,
after the ARN-based warping, the new pair ($I_a, I_b'$) would be well-aligned and become \emph{stationary pairs}.
In this scenario, the auto-mask~\cite{monodepth2} which was proposed to remove stationary points, are used in our framework for removing purely rotational points.
We implement this component in our baseline framework~\cite{bian2021ijcv},
and the effects of that is shown in \tabref{tab:ablation-am}.

\begin{table}[t]
\centering
    \caption{Single-view depth estimation results on NYUv2~\cite{silberman2012indoor}.
    }
    \label{tab:nyu}
    \scalebox{0.75}{
    \begin{tabular}{l c | c c c  | c c c}
     \toprule[1pt]
     \multirow{2}{*}{Methods} & \multirow{2}{*}{Supervision} & \multicolumn{3}{c|}{Error $\downarrow$} & \multicolumn{3}{c}{Accuracy $\uparrow$}  \\
     \cline{3-8}
      & & AbsRel & Log10 & RMS  & $\delta_1$ & $\delta_2$ & $\delta_3$ \\
     \hline
     Make3D~\cite{saxena2006learning} & \cmark &  0.349 & - & 1.214 & 0.447 & 0.745 & 0.897 \\
     Depth Transfer~\cite{karsch2014depth} & \cmark & 0.349 & 0.131 & 1.21 & - & - & - \\
     Liu~\etal~\cite{liu2014discrete} & \cmark &  0.335 & 0.127 & 1.06 & - & - & - \\ 
     Ladicky~\etal~\cite{ladicky2014pulling} & \cmark & - & - & - & 0.542 & 0.829 & 0.941 \\
     Li~\etal~\cite{li2015depth} & \cmark & 0.232 & 0.094 & 0.821 & 0.621 & 0.886 & 0.968 \\
     Roy~\etal~\cite{roy2016monocular} & \cmark & 0.187 & 0.078 & 0.744 & - & - & - \\
     Liu~\etal~\cite{liu2016learning} & \cmark & 0.213 & 0.087 & 0.759 & 0.650 & 0.906 & 0.976 \\
     Wang~\etal~\cite{wang2015towards} & \cmark & 0.220 & 0.094 & 0.745 & 0.605 & 0.890 & 0.970 \\
     Eigen~\etal~\cite{eigen2015predicting} & \cmark & 0.158 & - & 0.641 & 0.769 & 0.950 & 0.988 \\
     Chakrabarti~\etal~\cite{chakrabarti2016depth} & \cmark & 0.149 & - & 0.620 & 0.806 & 0.958 & 0.987 \\
     Laina~\etal~\cite{laina2016deeper} & \cmark & 0.127 & 0.055 & 0.573 & 0.811 & 0.953 & 0.988 \\
     Li~\etal~\cite{li2017two} & \cmark & 0.143 & 0.063 & 0.635 & 0.788 & 0.958 & 0.991 \\
     DORN~\cite{fu2018deep} & \cmark & 0.115 & 0.051 & 0.509 & 0.828 & 0.965 & 0.992 \\
     VNL~\cite{Yin2019enforcing} & \cmark & \textbf{0.108} & \textbf{0.048} & \textbf{0.416} & \textbf{0.875} & \textbf{0.976} & \textbf{0.994} \\
     \hline
     \hline
     MovingIndoor~\cite{Zhou_2019_ICCV} & \xmark & 0.208 & 0.086 & 0.712 & 0.674 & 0.900 & 0.968 \\
     TrainFlow~\cite{zhao2020towards} & \xmark & 0.189 & 0.079 & 0.686 & 0.701 & 0.912 & 0.978 \\
     Monodepth2~\cite{monodepth2} & \xmark & 0.176  &   0.074  &   0.639  &   0.734  &   0.937  &   0.983 \\
     P$^2$Net (3-frame)~\cite{IndoorSfMLearner} & \xmark & 0.159 & 0.068 & 0.599 & 0.772 & 0.942 & 0.984 \\
     P$^2$Net (5-frame)~\cite{IndoorSfMLearner} & \xmark & 0.147 & 0.062 & 0.553 & 0.801 & 0.951 & 0.987 \\
     \hline
     Baseline (SC-Depth~\cite{bian2021ijcv}) & \xmark  &   0.159  &   0.068  &   0.608  &   0.772  &   0.939  &   0.982   \\
     Ours-DP & \xmark &  0.143  &   0.060  &   0.538  &   0.812  &   0.951  &   0.986  \\
     Ours-ARN & \xmark & \textbf{0.138}  &   \textbf{0.059}  &   \textbf{0.532}  &   \textbf{0.820}  &   \textbf{0.956}  &   \textbf{0.989}  \\
     \bottomrule[1pt]
    \end{tabular}
    }
\end{table}

\begin{table}[t]
    \centering
    \caption{Unsupervised single-view depth estimation results on KITTI~\cite{Geiger2013IJRR}.
    }
    \label{tab:kitti}
    \scalebox{0.75}{
    \begin{tabular}{l  | c c c c | c c c}
     \toprule[1pt]
     \multirow{2}{*}{Methods} & \multicolumn{4}{c|}{Error $\downarrow$} & \multicolumn{3}{c}{Accuracy $\uparrow$}  \\
     \cline{2-8}
     & AbsRel & SqRel & RMS & RMSLog & $\delta_1$ & $\delta_2$ & $\delta_3$ \\
     \hline
     SfMLearner~\cite{zhou2017unsupervised} & 0.208 & 1.768 & 6.856 & 0.283 & 0.678 & 0.885 & 0.957 \\
     Vid2Depth~\cite{mahjourian2018unsupervised} & 0.163 & 1.240 & 6.220 & 0.250 & 0.762 & 0.916 & 0.968 \\
     DDAD~\cite{Wang2018CVPR} & 0.151 & 1.257 & 5.583 & 0.228 & 0.810 & 0.936 & 0.974 \\
     GeoNet~\cite{yin2018geonet} & 0.155 & 1.296 & 5.857 & 0.233 & 0.793 & 0.931 &  0.973 \\
     DF-Net~\cite{zou2018df} & 0.150 & 1.124 & 5.507 & 0.223 & 0.806 & 0.933 & 0.973 \\
     Struct2Depth~\cite{casser2019struct2depth} & 0.141 & 1.026 & 5.291 & 0.215 & 0.816 & 0.945 & 0.979 \\
     DW~\cite{gordon2019depth} & 0.128 & 0.959 & 5.230 & 0.212 & 0.845 & 0.947 &  0.976 \\
     GL-Net~\cite{chen2019self} & 0.135 & 1.070 & 5.230 & 0.210 & 0.841 & 0.948 & 0.980 \\
     CC~\cite{ranjan2019cc} & 0.140 & 1.070 & 5.326 & 0.217 & 0.826 & 0.941 & 0.975 \\
     EPC++~\cite{luo2019every} & 0.141 & 1.029 &  5.350 & 0.216 & 0.816 & 0.941 & 0.976 \\
     Monodepth2~\cite{monodepth2} & 0.115 & 0.882 & 4.701 & 0.190 & 0.879 & 0.961 & 0.982 \\
     TrainFlow~\cite{zhao2020towards} & 0.113 & \textbf{0.704} & 4.581 & 0.184 & 0.871 & 0.961 & \textbf{0.984} \\
     Insta-DM~\cite{lee2021learning} & 0.112 & 0.777 & 4.772 & 0.191 & 0.872 & 0.959 & 0.982 \\
     PackNet~\cite{packnet} & \textbf{0.107} & 0.802 & \textbf{4.538} & \textbf{0.186} & \textbf{0.889} & \textbf{0.962} & 0.981 \\
     \hline
     Baseline (SC-Depth~\cite{bian2021ijcv}) & 0.119 & \textbf{0.857} & 4.950 & 0.197 & 0.863 & 0.957 & \textbf{0.981} \\ 
     Ours & \textbf{0.118} & 0.861 & \textbf{4.803} & \textbf{0.193} & \textbf{0.866} & \textbf{0.958} & \textbf{0.981} \\
    \bottomrule[1pt]
    \end{tabular}
    }
\end{table}

\begin{table}[t]
\centering
    \caption{Zero-shot generalization results on ScanNet~\cite{dai2017scannet}.}
    \label{tab:scannet}
    \scalebox{0.78}{
    \begin{tabular}{l c | c c c  | c c c}
     \toprule[1pt]
     \multirow{2}{*}{Methods} & \multirow{2}{*}{Supervision} & \multicolumn{3}{c|}{Error $\downarrow$} & \multicolumn{3}{c}{Accuracy $\uparrow$}  \\
     \cline{3-8}
      & & AbsRel & Log10 & RMS  & $\delta_1$ & $\delta_2$ & $\delta_3$ \\
     \hline
     Laina~\etal~\cite{laina2016deeper} & \cmark &   0.141  &   0.059  &   0.339  &   0.811  &   0.958  &   0.990  \\ 
     VNL~\cite{Yin2019enforcing} & \cmark &  \textbf{0.123}  &   \textbf{0.052}  &   \textbf{0.306}  &   \textbf{0.848}  &   \textbf{0.964}  &   \textbf{0.991}  \\
     \hline
     \hline
     TrainFlow~\cite{zhao2020towards} & \xmark  &   0.179  &   0.076  &   0.415  &   0.726  &  0.927  &   0.980  \\
     SC-Depth~\cite{bian2021ijcv} & \xmark &  0.169  &   0.072  &   0.392  &   0.749  &   0.938  &   0.983  \\
     Ours & \xmark &  \textbf{0.156}  &   \textbf{0.066}  &   \textbf{0.361}  &   \textbf{0.781}  &   \textbf{0.947}  &   \textbf{0.987}   \\
     \toprule[1pt]
    \end{tabular}
    }
\end{table}

\paragraph{Can we reuse the PoseNet as ARN?}
It sounds possible because the ARN has a very similar network structure with the PoseNet.
Specifically, we can use only the rotation output of the PoseNet in the first stage for rectification and the full pose prediction in the second stage for image warping.
However, in practice, we are hard to make the network converge during training.
We conjecture that it depends on the model initialization and the knowledge conflicts between two steps.
More specifically, in the first stage it needs to learn coarse but big rotations, while in the second stage it needs to learn fine but small pose residuals.
Therefore, we suggest not sharing parameters between the ARN and PoseNet.

\section{Experiments}
\label{sec:experiment}

\subsection{Implementation Details}

We use SC-Depth~\cite{bian2021ijcv} as our baseline, which is publicly available.
We use the default hyper-parameters.
To be specific, we use $\alpha = 1$, $\beta = 0.1$, $\gamma = 0.5$ in \equref{eqn:loss}.
For training ARN, we use the weights $0.5$ for $L_{RT}$ and $0.1$ for $L_{RC}$.
We train models for $50$ epochs using the Adam~\cite{kingma2014adam} optimizer with the learning rate being $0.0001$. 
The batch size equals to $8$ for 3-frame inputs, and it is $16$ for pairwise inputs,
as the former is used as two pairs during training.
Besides, to demonstrate that our proposed pre-processing is universal to different methods,
we experiment with Monodepth2~\cite{monodepth2} (ResNet-18) in ablation studies,
where we use the default parameters and train models for $50$ epochs.

\subsection{Dataset and Metrics}

\paragraph{NYUv2~\cite{silberman2012indoor}.}
The dataset is composed of indoor video sequences recorded by a handheld Kinect RGB-D camera at $640 \times 480$ resolution.
The dataset contains 464 scenes taken from three cities.
We use the officially provided 654 densely labeled images for testing,
and use the rest $335$ sequences (no overlap with testing scenes) for training ($302$) and validation ($33$).
The raw training sequences contain $268K$ images.
It is first downsampled $10$ times to remove redundant frames.
We train on the pre-processed data (\ie, it generates $67K$ rectified pairs) and train directly on original data with ARN, respectively.
The images are resized to $320 \times 256$ resolution for training.

\paragraph{KITTI~\cite{Geiger2013IJRR}.}
The dataset contains driving videos in outdoor driving scenes,
and we use it to investigate whether the proposed ARN has an adverse impact on motion-simple data.
We use Eigen~\cite{eigen2014depth}'s splits and follow SC-Depth~\cite{bian2021ijcv} for training and evaluation.
The image is re-scaled to $832 \times 256$ resolution for training.

\paragraph{ScanNet~\cite{dai2017scannet}.}
The dataset provides RGB-D videos of $1513$ indoor scans, captured by handheld devices. 
We use officially released test set, which is originally used to evaluate the semantic labeling task, for evaluating the generalization of our trained model.
It contains total 2135 color images ($1296 \times 968$),
and corresponding ground truth depths ($640 \times 480$).
We re-scale the depth prediction to GT resolution for evaluation.

\paragraph{7-Scenes~\cite{shotton2013scene}.}
The dataset contains 7 indoor scenes, 
and each scene contains several video sequences ($500$-$1000$ frames per sequence), 
which are captured by a Kinect camera at $640 \times 480$ resolution.
We follow the official train/test split for each scene.
For testing, we simply extract the first image from every $10$ frames,
and for training, we fine-tune the model that is pre-trained on NYUv2 to demonstrate the universality of the proposed ARN.

\paragraph{Make3D~\cite{saxena2006learning}.}
Following previous unsupervised methods~\cite{zhou2017unsupervised, monodepth2}, we test the zero-shot generalization of our trained models on the Make3D dataset.
It contains $134$ in-the-wild images for testing.
However, caution should be taken, as the ground truth depth and input
images are not well aligned, causing potential evaluation issues.

\begin{table}[t]
\centering
    \caption{Camera pose estimation results on ScanNet~\cite{dai2017scannet}.}
    \label{tab:scannet-pose}
    \scalebox{1.0}{
    \begin{tabular}{l | c c c}
     \toprule[1pt]
     Method & rot (deg) & tr (deg) & tr (cm) \\
     \hline
    MovingIndoor~\cite{Zhou_2019_ICCV} & 1.96 & 39.17 & 1.40 \\
    Monodepth2~\cite{monodepth2} & 2.03 & 41.12 & 0.83 \\
    P$^2$Net~\cite{IndoorSfMLearner} & 1.86 & \textbf{35.11} & 0.89 \\
    Ours & \textbf{1.82} & 39.41 & \textbf{0.55} \\
     \toprule[1pt]
    \end{tabular}
    }
\end{table}

\begin{table}[t]
\centering
    \caption{Zero-shot generalization results on Make3D~\cite{saxena2006learning}.}
    \label{tab:make3d}
    \scalebox{0.9}{
    \begin{tabular}{l c | c c c }
     \toprule[1pt]
     Methods & Supervision & AbsRel & SqRel & RMS  \\
     \hline
     Karsch~\cite{karsch2014depth} & \cmark & 0.428 & 5.079 & 8.389 \\
     Liu~\cite{liu2014discrete} & \cmark & 0.475 & 6.562 & 10.05 \\
     Laina~\cite{laina2016deeper} & \cmark & \textbf{0.204} & \textbf{1.840} & \textbf{5.683} \\
     \hline
     \hline
     SfMLearner~\cite{zhou2017unsupervised} & \xmark & 0.383 & 5.321 & 10.470 \\
     DDVO~\cite{Wang2018CVPR} & \xmark & 0.387 & 4.720 & 8.090 \\
     Monodepth2~\cite{monodepth2} & \xmark & 0.322 & 3.589 & 7.417 \\
     Ours & \xmark & \textbf{0.305} & \textbf{2.869} & \textbf{7.320} \\
     \toprule[1pt]
    \end{tabular}
    }
\end{table}

\begin{table}[t]
\centering
    \caption{Fine-tuned results on 7-Scenes~\cite{shotton2013scene}.
    The model is pre-trained on NYUv2. We fine-tune models in each scene for $3$ epochs, which consumes less than $10$ minutes.
    }
    \label{tab:7-scene}
    \scalebox{0.9}{
    \begin{tabular}{l | c c |  c c}
     \toprule[1pt]
     \multirow{2}{*}{Scenes}  &\multicolumn{2}{c|}{No Fine-tuning} & \multicolumn{2}{c}{After Fine-tuning}  \\
     \cline{2-5}
     & AbsRel & Acc ($\delta_1$) & AbsRel & Acc ($\delta_1$) \\
     \hline
     Chess & 0.179 & 0.689 & \textbf{0.150} & \textbf{0.780} \\
     Fire  & 0.163 & 0.751 & \textbf{0.105} & \textbf{0.918} \\
     Heads & 0.171 & 0.746 & \textbf{0.143} & \textbf{0.833} \\
     Office& 0.146 & 0.799 & \textbf{0.128} & \textbf{0.855} \\
     Pumpkin& 0.120 & 0.841 & \textbf{0.097} & \textbf{0.922} \\
     RedKitchen & 0.136 & 0.822 & \textbf{0.124} & \textbf{0.853} \\
     Stairs & 0.143 & 0.794 & \textbf{0.134} & \textbf{0.823} \\
     \toprule[1pt]
    \end{tabular}
    }
\end{table}

\begin{table}[t]
\centering
    \caption{Effects of our proposed data processing (DP) on NYUv2~\cite{silberman2012indoor}.
    }
    \label{tab:ablation}
    \scalebox{0.82}{
    \begin{tabular}{l | c | c c c | c c c }
     \toprule[1pt]
     \multirow{2}{*}{Methods} & With & \multicolumn{3}{c|}{Error $\downarrow$} & \multicolumn{3}{c}{Accuracy $\uparrow$} \\
     \cline{3-8}
      & DP & AbsRel & Log10 & RMS  & $\delta_1$ & $\delta_2$ & $\delta_3$ \\
     \hline
     \multirow{2}{*}{SC-Depth~\cite{bian2021ijcv}} & \xmark  &   0.159  &   0.068  &   0.608  &   0.772  &   0.939  &   0.982   \\
     & \cmark  &  \textbf{0.143}  &   \textbf{0.060}  &   \textbf{0.538}  &   \textbf{0.812}  &   \textbf{0.951}  &   \textbf{0.986}  \\
     \hline
     \multirow{2}{*}{Monodepth2~\cite{monodepth2}} & 
     \xmark &
     0.176  &   0.074  &   0.639  &   0.734  &   0.937  &   0.983 \\  
     & \cmark &
     \textbf{0.151}  &  \textbf{ 0.064}  &   \textbf{0.559}  &   \textbf{0.795}  &  \textbf{0.947}  &   \textbf{0.985}
     \\
     \toprule[1pt]
    \end{tabular}
    }
\end{table}

\paragraph{Evaluation metrics.}
Following previous methods~\cite{liu2016learning,laina2016deeper,fu2018deep, Yin2019enforcing}, we use the mean absolute relative error (AbsRel),
mean log10 error (Log10), root mean squared error (RMS),
root mean squared log error (RMSLog),
and the accuracy under threshold ($\delta_i$ $<$ $1.25^i$, $i = 1, 2, 3$).
As unsupervised methods cannot recover the metric scale, we multiply the predicted depth maps by a scalar that matches the median with that of the ground truth,
as in~\cite{zhou2017unsupervised,bian2021ijcv,monodepth2}.
The predicted depths are capped at $80m$ in KITTI, $70m$ in Make3D, and $10m$ in all indoor datasets.


\begin{figure*}[t]
\centering
\begin{tabular}{c c}
   \includegraphics[width=0.45\linewidth]{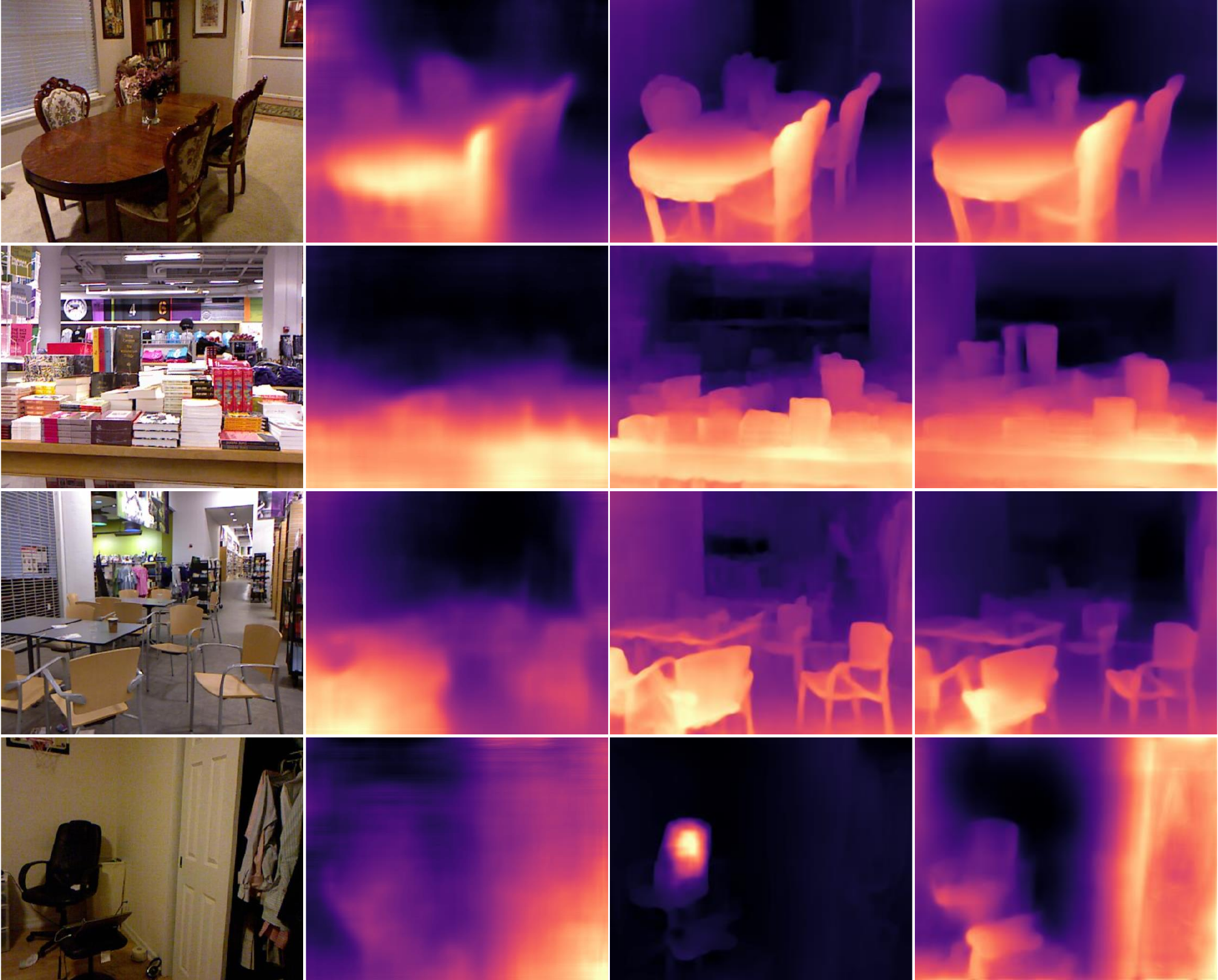} & \includegraphics[width=0.45\linewidth]{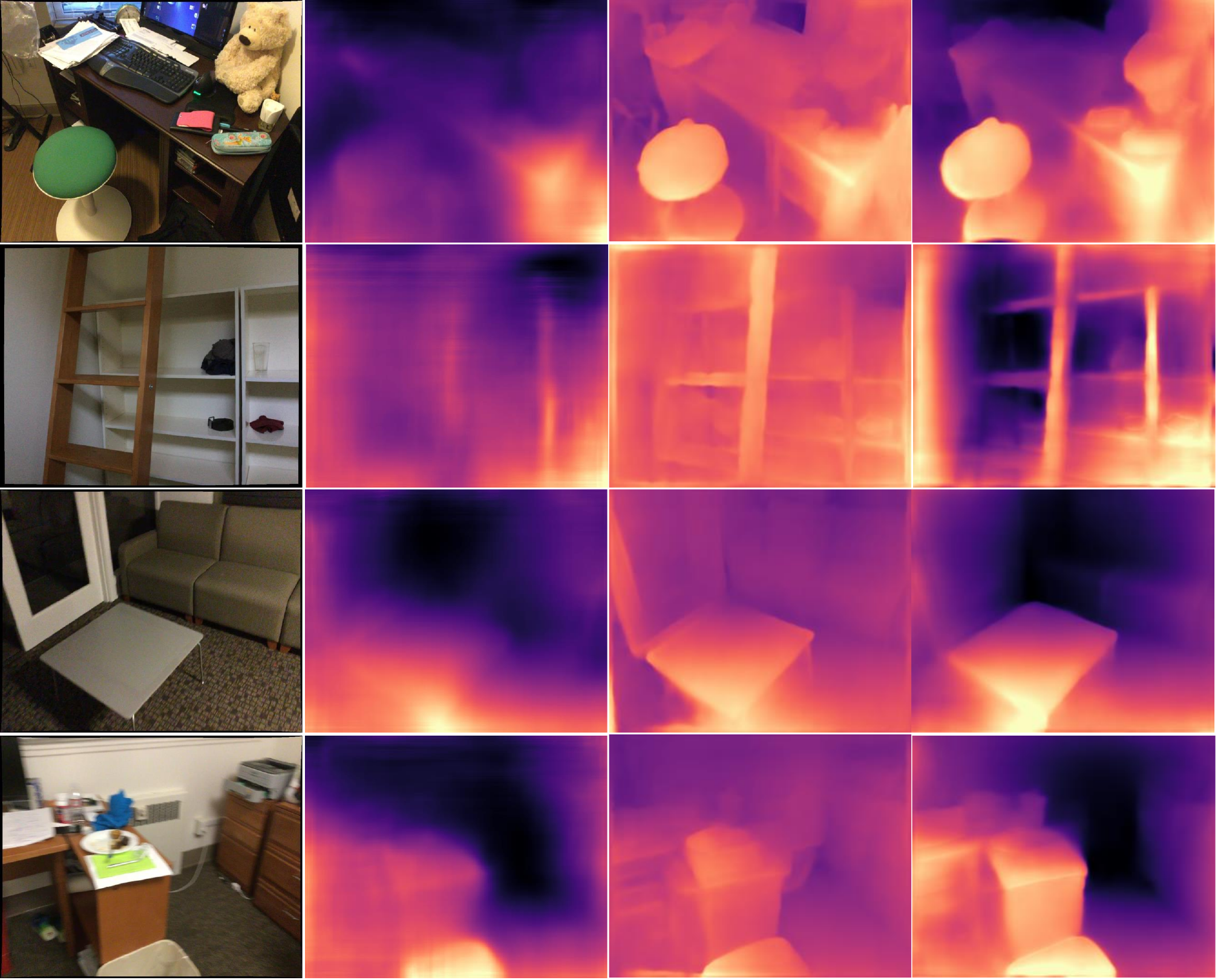} \\
   (a) NYUv2 & (b) ScanNet \\
   \includegraphics[width=0.45\linewidth]{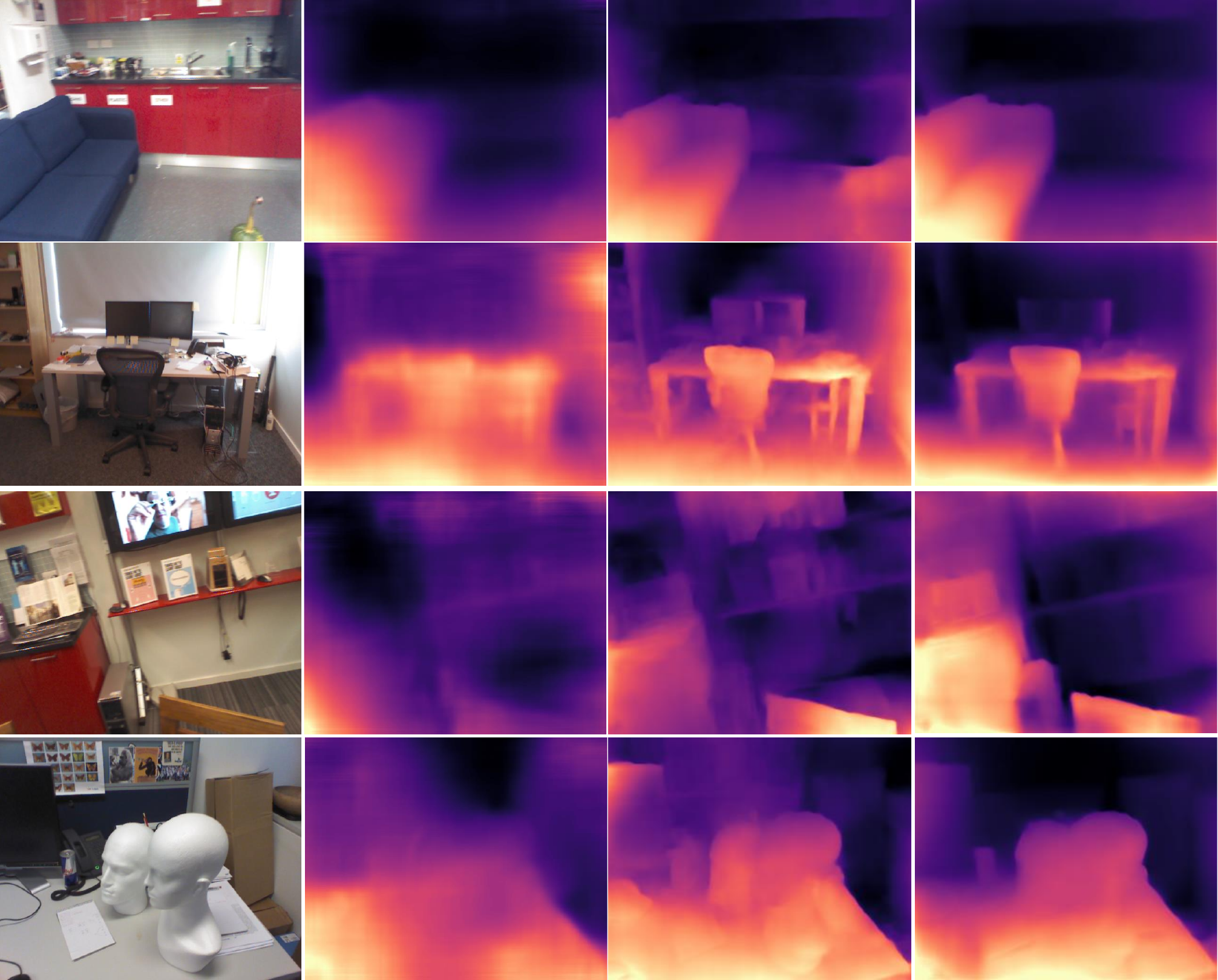} &
   \includegraphics[width=0.45\linewidth]{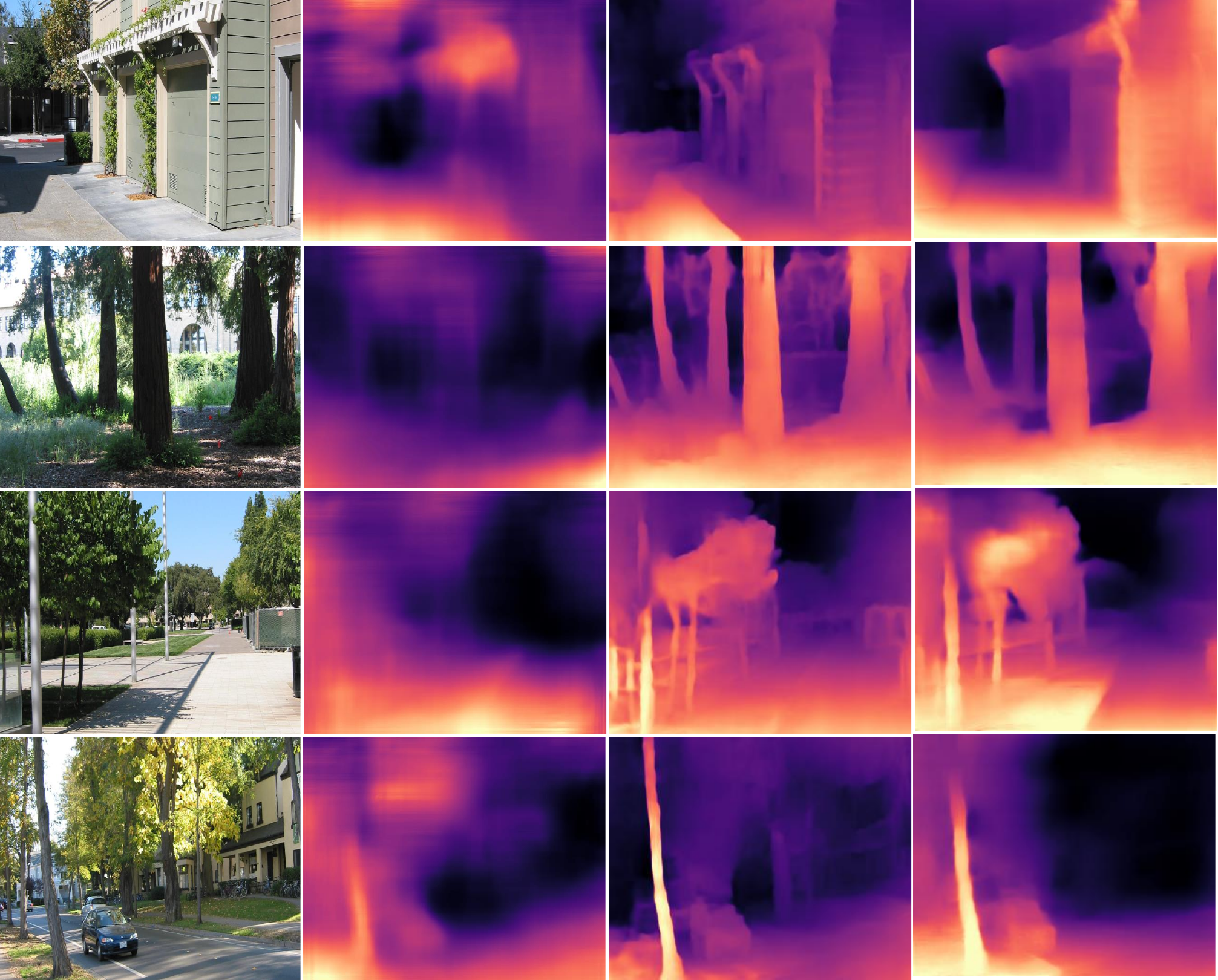} \\ 
    (c) 7-Scenes & (d) Make3D \\
\end{tabular}
\caption{Qualitative results. Left to right: RGB, TrainFlow~\cite{zhao2020towards}, Monodepth2~\cite{monodepth2}, and Ours. The models are trained on NYUv2~\cite{silberman2012indoor}.}
\label{fig:show}
\end{figure*}

\subsection{Results}\label{sec:results}

\begin{figure*}[t]
\centering
\includegraphics[width=0.95\linewidth]{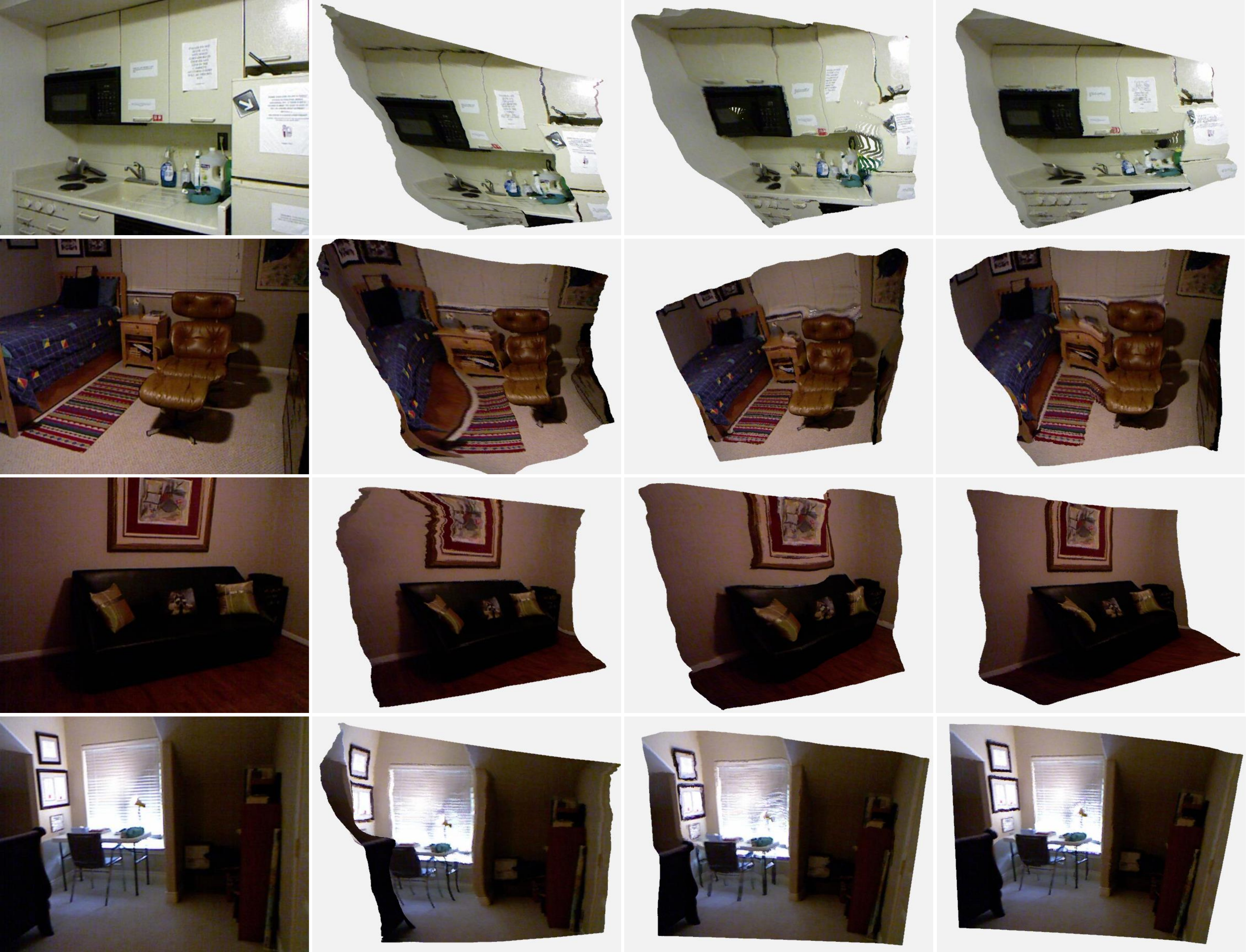}
\caption{Visualisation of 3D point clouds on NYUv2~\cite{silberman2012indoor}. Left to right: RGB, TrainFlow~\cite{zhao2020towards}, Monodepth2~\cite{monodepth2}, and Ours.}
\label{fig:pcd-vis}
\end{figure*}

\paragraph{Results on NYUv2.}
\tabref{tab:nyu} shows the single-view depth estimation results on NYUv2~\cite{silberman2012indoor}.
It shows that our method clearly outperforms previous unsupervised methods~\cite{zhao2020towards, monodepth2, IndoorSfMLearner}.
See qualitative results of depth estimation in \figref{fig:show} and the visualization of 3D point clouds in \figref{fig:pcd-vis}. 
Besides, compared with SC-Depth~\cite{bian2021ijcv}, which is our baseline method,
the higher performance demonstrates the efficacy of our proposed DP and ARN.
Note that NYUv2 dataset is so challenging that many previous unsupervised methods such as GeoNet~\cite{yin2018geonet} fail to get reasonable results, as discussed in \cite{Zhou_2019_ICCV}.
We find that Monodepth2~\cite{monodepth2} sometimes also meets this issue,
and we report its best result of multiple runs.
Moreover, our method outperforms a series of fully supervised methods~\cite{liu2016learning, saxena2006learning, karsch2014depth, liu2014discrete, ladicky2014pulling, li2015depth, roy2016monocular, wang2015towards, eigen2015predicting, chakrabarti2016depth, li2017two}.
However, there still has a gap to the state-of-the-art supervised method~\cite{Yin2019enforcing}.

\paragraph{Results on KITTI}.
Other than indoor datasets, we also report the results on the KITTI outdoor driving dataset~\cite{Geiger2013IJRR}.
Note that in the driving dataset, the image rotation is very small because most image pairs have a simple forward motion,
and in this scenario our proposed rotation rectification is not necessary.
Therefore, we do not run the data processing (DP) and only evaluate the effects of our proposed ARN.
The results are summarized in \tabref{tab:kitti}.
It shows that the performance of our method is slightly lower than the state-of-the-art methods~\cite{monodepth2, zhao2020towards, packnet}.
However, compared with SC-Depth~\cite{bian2021ijcv}, which is our baseline method,
the proposed ARN module can lead to a minor improvement.
It demonstrates that the proposed ARN has no adverse impact when training on motion-simple datasets.

\paragraph{Generalization results on ScanNet.}
\tabref{tab:scannet} shows the zero-shot generalization results on ScanNet~\cite{dai2017scannet},
where all models are trained on NYUv2~\cite{silberman2012indoor}.
The results demonstrate that our trained models generalize well to new dataset.
See qualitative results in \figref{fig:show}.
Besides, following \cite{IndoorSfMLearner}, we provide the pose estimation results on ScanNet dataset,
where $2000$ image pairs from diverse scenes are selected by \cite{teed2018deepv2d}.
The results demonstrate that our method outperforms other unsupervised alternatives.

\paragraph{Generalization results on Make3D.}
\tabref{tab:make3d} shows the zero-shot generalization results on Make3D~\cite{saxena2006learning}.
Note that it is very challenging because our models are trained on indoor NYUv2 dataset but tested on in-the-wild outdoor images.
Here, even though other unsupervised methods are trained on KITTI outdoor dataset,
in which the images are arguably more similar to Make3D images,
the results show that our indoor trained models outperform other unsupervised outdoor trained models.

\paragraph{Fine-tuned results on 7-Scenes.}
The nature of unsupervised learning makes our method easy to be fine-tuned in new datasets, \ie, we can fine-tune models there without the requirement for the ground-truth labels.
\tabref{tab:7-scene} shows the quantitative fine-tuned results on 7-Scenes~\cite{shotton2013scene},
where the models were pre-trained on NYUv2~\cite{silberman2012indoor}.
The results shows that our model not only generalizes well to new datasets,
but also a quick fine-tuning can improve the performance significantly.
This has important implications to real-world applications.
Also, the results demonstrate that our proposed ARN performs well in different scenarios.

\paragraph{Timing.}
It takes $28$ hours to train the models for $50$ epochs on original NYUv2 dataset and $25$ hours on the rectified data,
measured in a single 16GB NVIDIA V100 GPU.
It takes $44$ hours when training with the proposed ARN.
The inference speed is about $210$ FPS in a NVIDIA RTX 2080 GPU,
where images are resized to $320 \times 256$ before feeding to the network.



\subsection{Ablation studies}

\begin{figure}[t]
\centering
\includegraphics[width=0.95\linewidth]{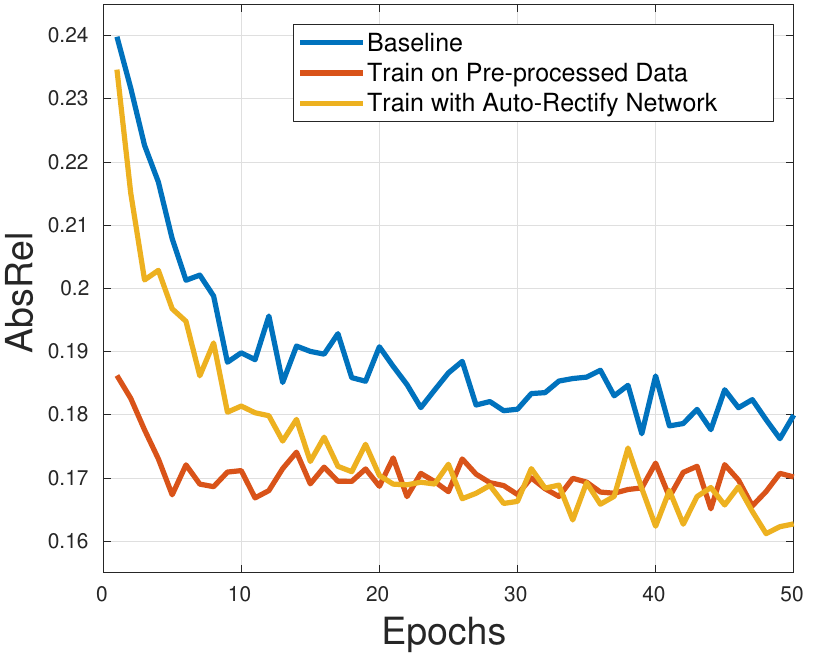}
\caption{Validation loss during training on NYUv2~\cite{silberman2012indoor}.}
\label{fig:validation-loss}
\end{figure}

\paragraph{Effects of the proposed DP.}
\tabref{tab:ablation} shows the ablation study results on NYUv2.
For both SC-Depth~\cite{bian2021ijcv} and Monodepth2~\cite{monodepth2} frameworks,
training on our pre-processed data leads to significantly improved results than using original dataset.
It demonstrates that the proposed data processing method is independent to the method chosen,
and our motivation of removing rotation is correct.
Moreover, we visualize the quantitative learning curves in \figref{fig:validation-loss} for more detailed comparison.

\paragraph{Effects of the proposed ARN.}
The results in \tabref{tab:nyu} show that the proposed ARN improves the depth estimation accuracy significantly on NYUv2.
Besides, the results in \tabref{tab:kitti} show that the performance improvement by ARN is minor on KITTI,
since the camera rotation in driving scenes is almost small.
As shown in \figref{fig:pose-statistics}(a), the ego-motion in NYUv2 is more complex than in KITTI. 
This indicates that using ARN is beneficial to 
training on motion-complex data (see \figref{fig:validation-loss}),
while it has less impact when training on motion-simple data.

\paragraph{Effects of the proposed losses.}
\tabref{tab:ablation-loss} shows the ablation study results of the proposed loss functions on NYUv2.
Note that the ARN could converge well even without applying the proposed loss functions during training,
since it can get supervision signals from the photometric loss and geometry consistency loss, contributed to the differentiable warping.
This is very important because it makes our system really end-to-end trainable,
leading to improved performance against the two-step learning---our data preprocessing solution.
Besides, the results show that the proposed $L_{RT}$ and $L_{RC}$ can effectively regularize the network during training and lead to a higher performance.
Moreover, \tabref{tab:ablation-am} shows the effects of ImageNet pretraining and Auto-Mask, which is used to remove purely rotational motions.


\begin{table}[t]
\centering
    \caption{Effects of the proposed loss functions on NYUv2~\cite{silberman2012indoor}.}
    \label{tab:ablation-loss}
    \scalebox{0.85}{
    \begin{tabular}{l | c | c c c | c c c }
     \toprule[1pt]
     With  & With & \multicolumn{3}{c|}{Error $\downarrow$} & \multicolumn{3}{c}{Accuracy $\uparrow$} \\
     \cline{3-8}
     $L_{RT}$ & $L_{RC}$ & AbsRel & Log10 & RMS  & $\delta_1$ & $\delta_2$ & $\delta_3$ \\
     \hline
     \xmark & \xmark &   0.150  &   0.064  &   0.564  &   0.797  &   0.946  &   0.985  \\
     \cmark & \xmark &  0.145  &   0.062  &   0.560  &   0.802  &   0.948  &   0.987 \\
     \xmark & \cmark &   0.148  &   0.063  &   0.562  &   0.798  &   0.950  &   0.987 \\
     \cmark & \cmark &  \textbf{0.138}  &   \textbf{0.059}  &   \textbf{0.532}  &   \textbf{0.820}  &   \textbf{0.956}  &   \textbf{0.989} \\
     \toprule[1pt]
    \end{tabular}
    }
\end{table}

\begin{table}[t]
\centering
    \caption{Effects of Auto-Mask (AM) and ImageNet Pretrain (IP) on NYUv2~\cite{silberman2012indoor}.}
    \label{tab:ablation-am}
    \scalebox{0.95}{
    \begin{tabular}{l | c | c c c | c c c }
     \toprule[1pt]
     With  & With & \multicolumn{3}{c|}{Error $\downarrow$} & \multicolumn{3}{c}{Accuracy $\uparrow$} \\
     \cline{3-8}
     AM & IP & AbsRel & Log10 & RMS  & $\delta_1$ & $\delta_2$ & $\delta_3$ \\
     \hline
     \cmark & \xmark &  0.164  &   0.069  &   0.596  &   0.768  &   0.937  &   0.982  \\
     \xmark & \cmark &  0.153  &   0.065  &   0.571  &   0.790  &   0.945  &   0.986  \\
     \cmark & \cmark &  \textbf{0.138}  &   \textbf{0.059}  &   \textbf{0.532}  &   \textbf{0.820}  &   \textbf{0.956}  &   \textbf{0.989} \\
     \toprule[1pt]
    \end{tabular}
    }
\end{table}

\begin{figure}[t]
\centering
\begin{tabular}{cc}
\includegraphics[width=0.45\linewidth]{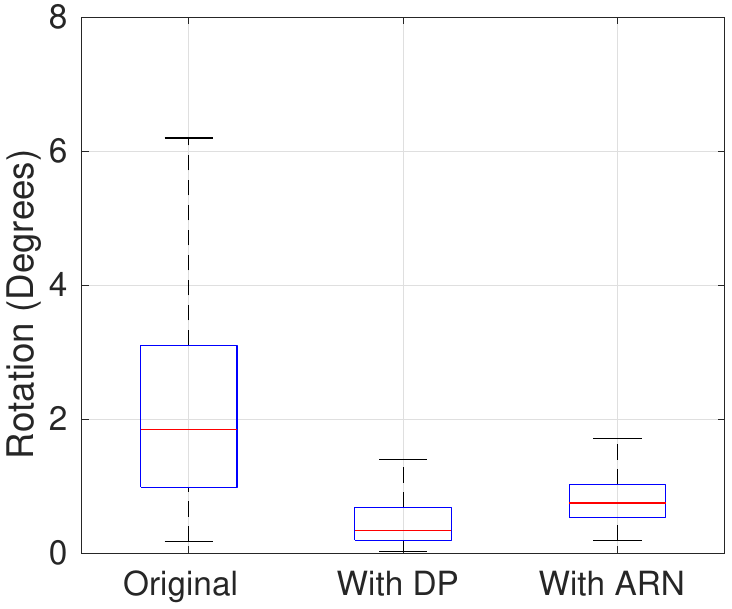} &
\includegraphics[width=0.45\linewidth]{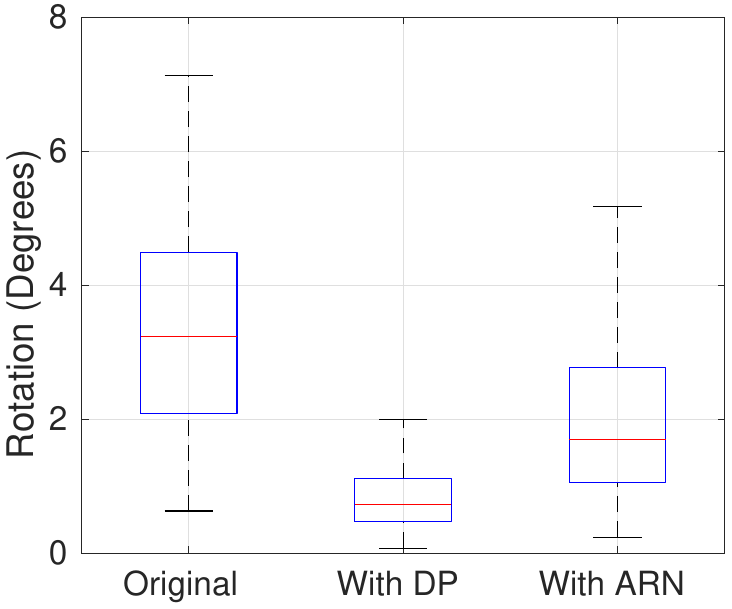} \\
NYUv2 & 7-Scenes \\
\end{tabular}
\caption{Effects of proposed rectification methods. We test the data pre-processing (DP) and ARN on the validation sequence.
The ARN model is trained on NYUv2~\cite{silberman2012indoor}.}
\label{fig:arn-rot}
\end{figure}

\paragraph{Rotation removal by ARN.}
\figref{fig:arn-rot} shows that the proposed ARN can effectively reduce the relative image rotations,
where the model is trained on NYUv2 without fine-tuning on 7-Scenes.
However, when compared with traditional geometry based methods,
the accuracy is lower, and the generalization is worse.
Besides, \figref{fig:arn-vis} shows several ARN warped results during training,
and \figref{fig:validation-loss} demonstrates that this is effective for improving depth learning.
Therefore, we conclude that ARN is effective for assisting the depth learning,
while not comparable to geometry-based methods in terms of rotation estimation.

\section{Conclusion}
In this paper, we analyze the impact of ego-motion 
on the indoor unsupervised depth learning problem.
Our analysis shows that (i) rotational motions dominate the camera motions in videos taken by handheld devices,
and (ii) rotation behaves as noises while translation contributes effective signals to training.
Based on the analysis, we propose a data pre-processing method to rectify the training image pairs.
The results show that training on rectified image pairs can effectively improve the performance.
Besides, we propose an Auto-Rectify Network with novel loss functions,
which are integrated into the existing self-supervised learning framework for end-to-end learning.
The comprehensive evaluation results in different datasets demonstrate the efficacy and universality of our proposed methods.

\ifCLASSOPTIONcompsoc
  \section*{Acknowledgments}
\else
  \section*{Acknowledgment}
\fi
This work was in part supported by the Australian Centre of Excellence for Robotic Vision CE140100016, 
and the ARC Laureate Fellowship FL130100102 to Prof. Ian Reid. 
We thank anonymous reviewers for their valuable suggestions.

\ifCLASSOPTIONcaptionsoff
  \newpage
\fi

\bibliographystyle{IEEEtran}
\bibliography{reference}

\newcommand{\AddPhoto}[1]{\includegraphics%
[width=1in,keepaspectratio]{#1}}

\begin{IEEEbiography}[\AddPhoto{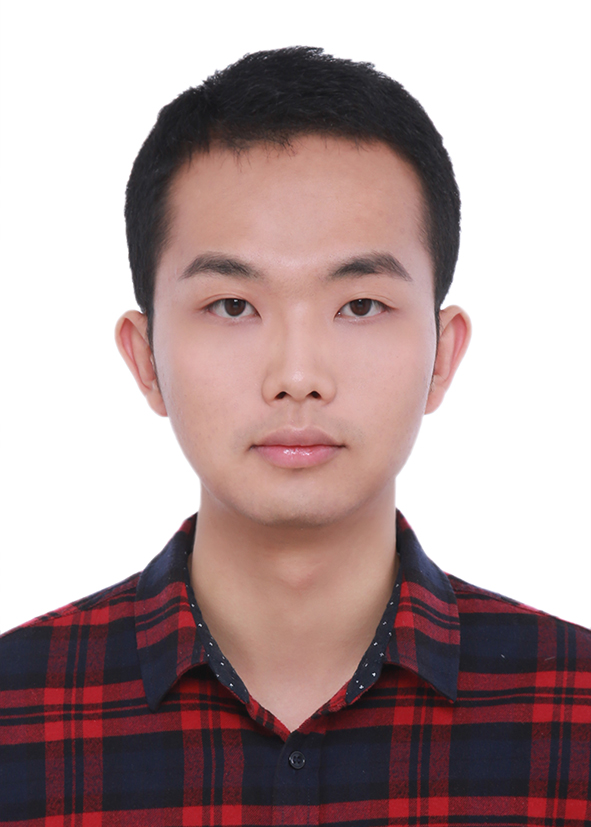}]{Jia-Wang Bian}
is a PhD candidate at the University of Adelaide and an Associated PhD researcher with the Australian Centre for Robotic Vision (ACRV). He is advised by Prof. Ian Reid and Prof. Chunhua Shen. His research interests lie in the field of computer vision, deep learning, and robotics. Jiawang received his B.Eng. degree from Nankai University, where he was advised by Prof. Ming-Ming Cheng. He was a research assistant at the Singapore University of Technology and Design (SUTD). Also, Jiawang did research intern jobs in many companies, including Advanced Digital Sciences Center (ADSC), Huawei, TuSimple, Amazon, and Meta.
\end{IEEEbiography} 
\vspace{-0.5cm}

\begin{IEEEbiography}[\AddPhoto{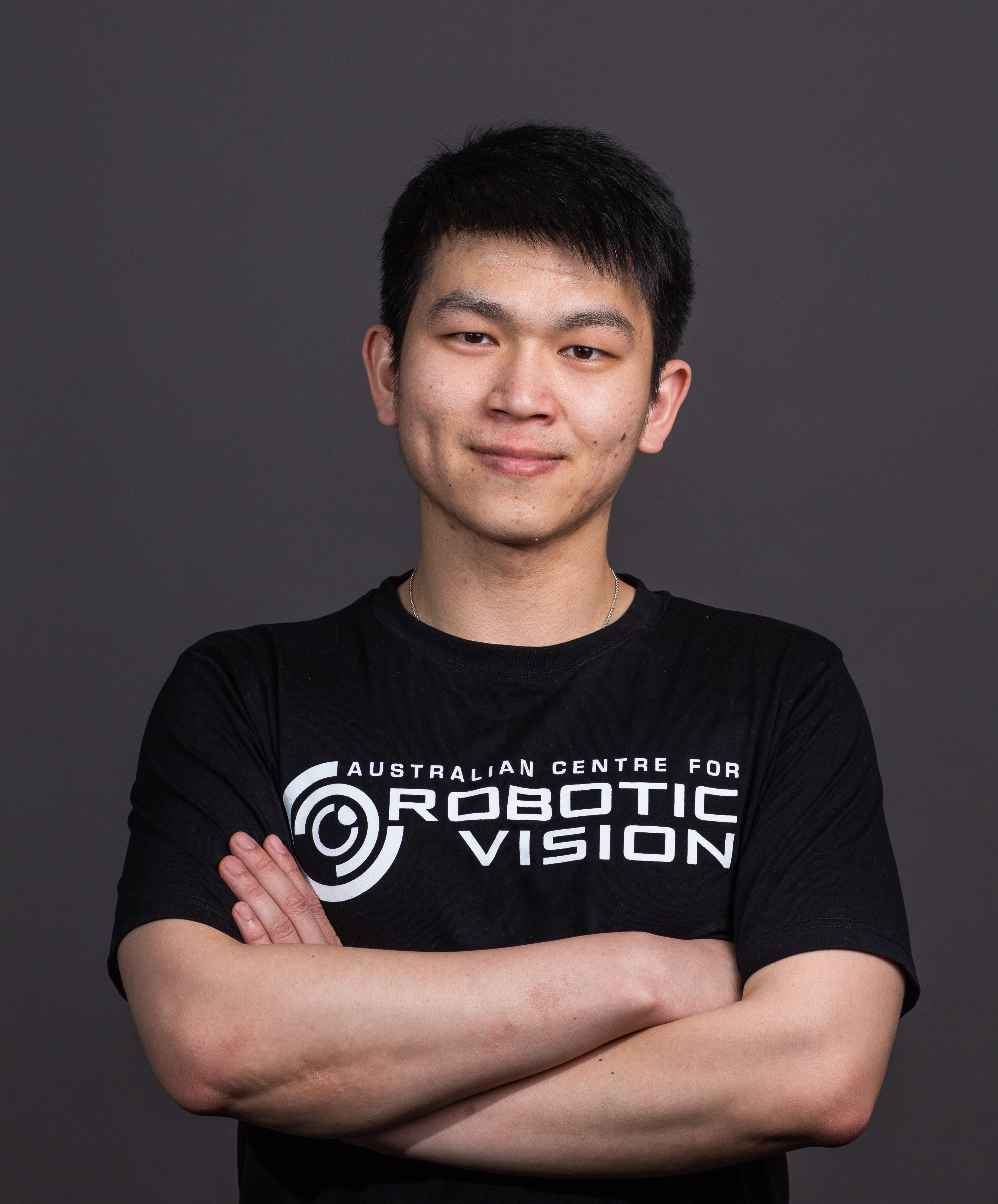}]{Huangying Zhan}
is a Postdoctoral Researcher at the University of Adelaide, where he is working with Prof. Ian Reid @ Uni.Adelaide and Dr. Hamid Rezatofighi @ Monash University. He is associated with The Australian Institute for Machine Learning (AIML) @ UniAdelaide and Vision \& Learning for Autonomous AI Lab(VL4AI) @ Monash University. His research interests include deep learning and its application in robotic vision and visual navigation.
Piror to that, he received his Ph.D. degree from the University of Adelaide and he was affiliated with the Australian Centre for Robotic Vision, where he was advised by Prof. Ian Reid and Prof. Gustavo Carneiro. Previously, he received my B.Eng degree in Electronic Engineering (first class honors) from The Chinese University of Hong Kong (CUHK), where he was advised by Prof. Xiaogang Wang. Also, he was a visiting student in the Unmanned Systems Research Group at The National University of Singapore, where he worked with Prof. Ben M. Chen.
\end{IEEEbiography} 
\vspace{-0.5cm}

\begin{IEEEbiography}[\AddPhoto{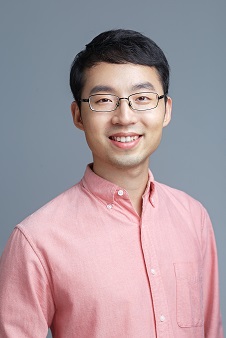}]{Naiyan Wang}
is currently the chief scientist of TuSimple. he leads the algorithm research group in the Beijing branch. Before this, he got his PhD degree from CSE department, HongKong University of Science and Technology in 2015. His supervisor is Prof. Dit-Yan Yeung. He got his BS degree from Zhejiang University, 2011 under the supervision of Prof. Zhihua Zhang.
His research interest focuses on applying statistical computational model to real problems in computer vision and data mining. Currently, He mainly works on the vision based perception and localization part of autonomous driving. Especially He integrates and improves the cutting-edge technologies in academia, and makes them work properly in the autonomous truck.
\end{IEEEbiography} 
\vspace{-0.5cm}

\begin{IEEEbiography}[\AddPhoto{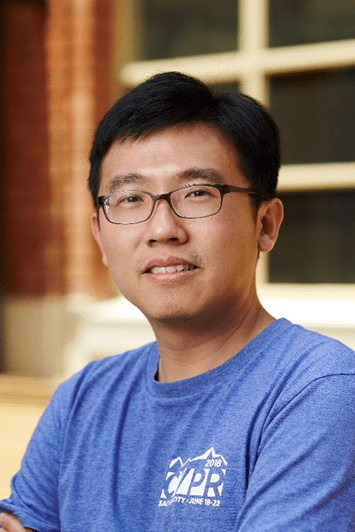}]{Tat-Jun Chin} 
received his PhD in Computer Systems Engineering from Monash University in 2007, which was supported by the Endeavour Australia-Asia Award, and a Bachelor in Mechatronics Engineering from Universiti Teknologi Malaysia in 2004, where we won the Vice Chancellor’s Award. He currently holds the SmartSat CRC Professorial Chair of Sentient Satellites at The University of Adelaide. He is also the Director of Machine Learning for Space at The Australian Institute for Machine Learning. Tat-Jun’s research interest lies in optimisation for computer vision and machine learning, and their application to robotic vision, space and smart cities. He has published more than 100 research articles on the subject, and has won several awards for his research, including a CVPR award (2015), a BMVC award (2018), Best of ECCV (2018), two DST Awards (2015, 2017) and IAPR Award (2019).
\end{IEEEbiography} 
\vspace{-0.5cm}

%
%

\begin{IEEEbiographynophoto}
{Chunhua Shen} is a Professor of Computer Science at the
University of Adelaide, Australia.
\end{IEEEbiographynophoto}
\vspace{-0.5cm}

\begin{IEEEbiography}[\AddPhoto{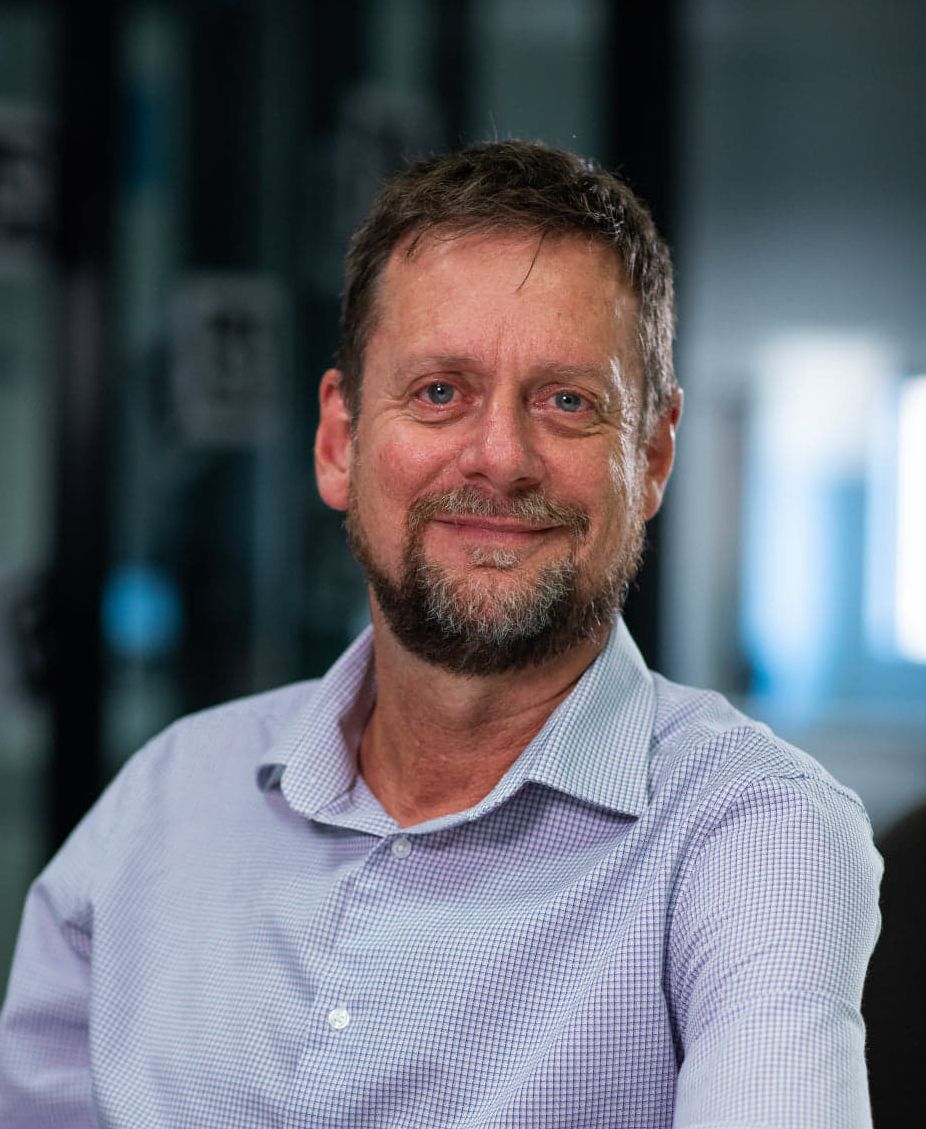}]{Ian Reid} 
is the Head of the School of Computer Science at the University of Adelaide, and the senior researcher at the Australian Institute for Machine Learning.
He is a Fellow of Academy of Technological Sciences and Engineering and held an ARC Australian Laureate Fellowship there 2013-18. He is also Deputy Director of the ARC Centre of Excellence in Robotic Vision, in which capacity he also co-chaired sub-committees of the National Roadmap for Robotics produced and published by the Centre in 2018.
Between 2000 and 2012 he was a Professor of Engineering Science at the University of Oxford. where, together with long-time colleague Prof. David Murray, he ran the Active Vision Group which is part of the wider Robotics Research Group.
He received a BSc in Computer Science and Mathematics with first class honours from University of Western Australia in 1987 and was awarded a Rhodes Scholarship in 1988 to study at the University of Oxford, where he obtained a D.Phil. in 1992.
\end{IEEEbiography} 

\end{document}